%% file: main.tex
\documentclass[11pt]{article}

\usepackage[utf8]{inputenc}
\usepackage[T1]{fontenc}
\usepackage{lmodern}
\usepackage[margin=1in]{geometry}
\usepackage{microtype}
\usepackage{graphicx}
\usepackage{booktabs}
\usepackage{amsmath}
\usepackage{amssymb}
\usepackage{caption}
\usepackage{listings}
\usepackage{xcolor}
\usepackage{placeins}
\usepackage[numbers,sort&compress]{natbib}
\usepackage[hidelinks]{hyperref}
\usepackage{xurl}

\hypersetup{
  pdftitle={Scientific Agent Skills: A Library of Procedural Knowledge for Research Agents},
  pdfauthor={Timothy Kassis, Vinayak Agarwal, Yuhuan He, Darshil Patel, Aubrey M. Brueckner}
}

\renewcommand{\topfraction}{0.8}

\renewcommand{\textfraction}{0.15}
\renewcommand{\floatpagefraction}{0.7}

\definecolor{codebg}{HTML}{F6F6F3}
\lstdefinestyle{skillmd}{
  basicstyle=\ttfamily\footnotesize,
  backgroundcolor=\color{codebg},
  frame=single,
  rulecolor=\color{lightgray},
  framesep=5pt,
  breaklines=true,
  columns=fullflexible,
  keepspaces=true,
  showstringspaces=false,
  xleftmargin=0pt,
}

\input{numbers}

\title{Scientific Agent Skills: A Library of Procedural Knowledge for
Research Agents}

\author{%
  Timothy Kassis\thanks{Correspondence: \texttt{timothy.kassis@k-dense.ai}},
  Vinayak Agarwal, Yuhuan He, Darshil Patel, Aubrey M. Brueckner \\
  K-Dense, Inc.
}

\date{\today}

\begin{document}
\maketitle

% Graphical abstract. Deliberately unnumbered and outside the `figure`
% environment: it is a frontispiece, not a figure the text refers to, and
% numbering it would renumber every cross-reference in the body. Generated by
% `uv run python -m sas_paper.graphical_abstract`, which holds the prompt.
% The two gaps around the frontispiece are tightened so the abstract's last
% line does not butt against the footnote rule. Page one is a fixed budget:
% the frontispiece and the whole abstract must fit on it.
\vspace{-1.2em}
\begin{center}
  \includegraphics[width=0.92\textwidth]{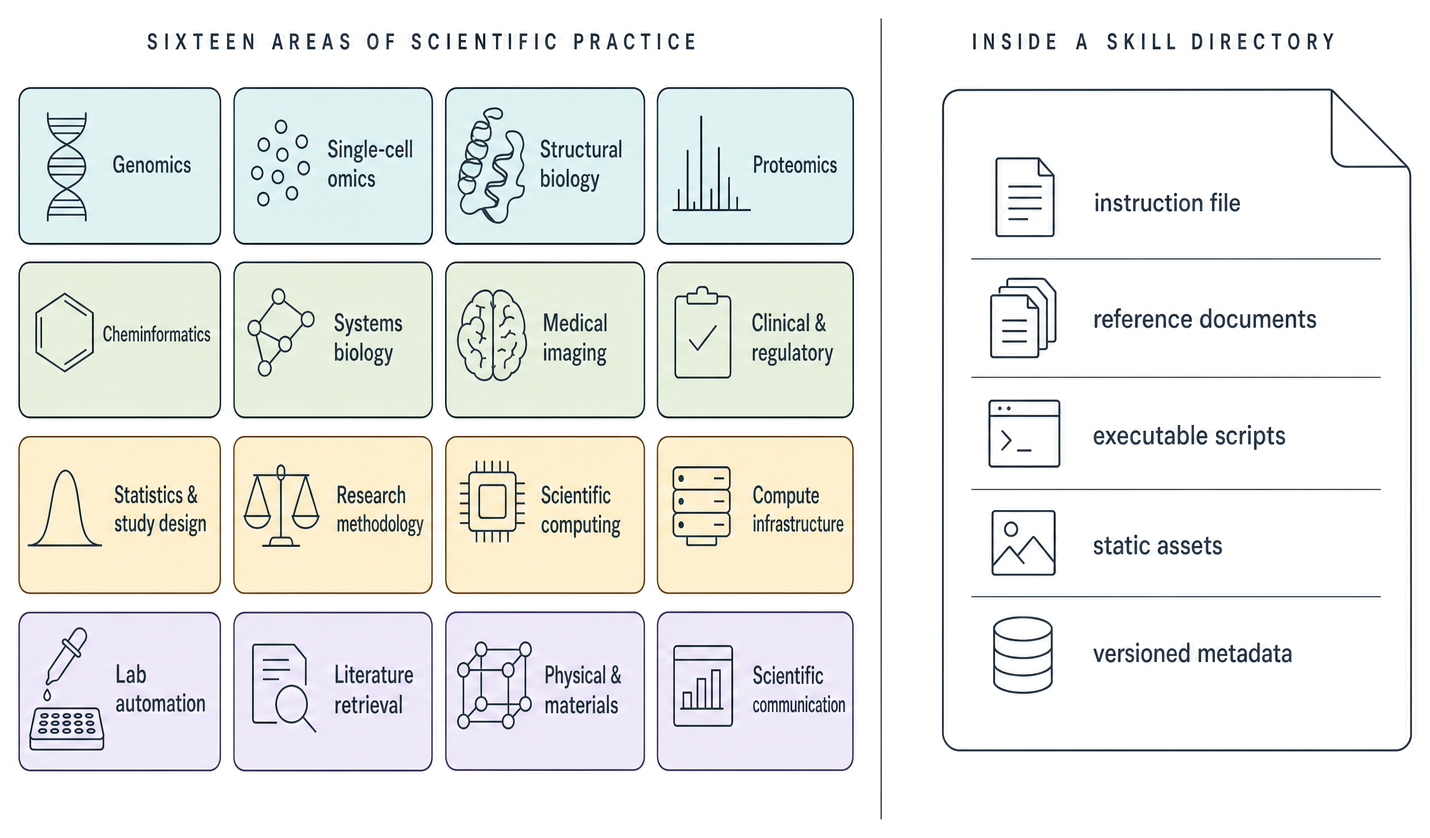}
\end{center}
\vspace{-1.2em}

\begin{abstract}
A language-model agent asked to analyse an experiment will usually return
working code.
Whether the analysis is defensible is a different question.
A defensible analysis depends on procedural choices: which test the field
accepts, which identifier namespace is authoritative, and which caveats must
accompany a result.
We present Scientific Agent Skills, an open library of \NumSkills{} such
procedures in \NumCategories{} areas of practice, including genomics,
cheminformatics, medical imaging, study design and scientific communication.
Each skill is a directory built around a versioned, human-readable instruction
file.
An agent loads the file only when a task calls for it; the directory often also
contains reference material and runnable scripts.
We report no task-level evaluation and no host selection rate.
We measure two properties of the documentation corpus: the always-resident
descriptions of all \NumSkills{} skills cost \ResidentFracWindow\% of a
\WindowTokens{}-token window, and the median documented workflow fits within
\WorkflowMedianInstructionPct\% of it, although \WorkflowOverWindow{} of
\NumDocWorkflows{} would overflow if every reference file were loaded.
Openly licensed and available at
\url{https://github.com/K-Dense-AI/scientific-agent-skills}.
\end{abstract}

% Page one is title + frontispiece + abstract only; the body starts on page 2.
% The check is that page 2 begins with `Introduction'.
\clearpage

\section{Introduction}

A researcher who asks a coding agent to analyse an RNA-seq experiment will
usually get working code.
Whether they get a defensible analysis is a different question.
For example, the agent may compare groups without correcting for multiple
testing, count a hundred cells from each of three mice as three hundred
independent replicates, process every treated sample on one day and every
control on another, or treat a BED interval's zero-based start as one-based.
Each mistake can produce plausible code while invalidating the analysis under
the stated conditions.
The relevant conventions are documented: multiple-testing correction is
standard~\citep{benjamini1995controlling}, pseudoreplication recurs in
published neuroscience~\citep{lazic2010pseudoreplication}, technical
confounding affects high-throughput analyses~\citep{leek2010tackling}, BED and
related formats use explicit coordinate
conventions~\citep{ucsc2026coordinates}, and spreadsheet conversion of gene
symbols has propagated into published work~\citep{ziemann2016gene}.
An evaluation of agents on real scientific requests found that scientific
accuracy trailed communication within every model and that overclaiming was the
most common failure tag~\citep{brueckner2026kbench}.

The recurring problem is procedural: agents need field-specific conventions
that specify which test applies, which identifier namespace is authoritative,
and which caveats must accompany a result.
That knowledge is distributed across package documentation, reporting
guidelines, standards, and the tacit practice of individual fields.
Rediscovering it for every task is costly, so it can be recorded once.
Scientific communities already record such knowledge for people.
Reporting guidelines such as ARRIVE~\citep{percie2020arrive} and
MIQE~\citep{bustin2009miqe} exist because the same procedural mistakes recur
across the published literature~\citep{makin2019ten}.

Agent Skills~\citep{agentskills2026spec} are a convention for recording this
procedural knowledge for agents.
A skill is a directory containing a Markdown instruction file with a short YAML
header, optionally accompanied by reference documents, executable scripts and
static assets.
A host keeps only each skill's name and short description in context and reads
the full file when a task appears to call for it (Section~\ref{sec:what}).
The convention specifies no runtime, API or service, so a skill is portable
across hosts and reviewable as text.

This paper describes Scientific Agent Skills~\citep{kdense2026skills}, an
openly licensed library of \NumSkills{} skills that we build and maintain,
covering research workflows in biology, chemistry, medicine, the physical
sciences and scientific communication.
Each mistake in the opening paragraph is answered by a rule in one of these
skills: \texttt{statistical-analysis} requires the agent to ``say which
correction was used''; \texttt{experimental-design} states that ``3 mice with
100 cells each is n = 3 (mice), not n = 300 (cells)''; \texttt{bulk-rnaseq}
warns that when treated and control samples were processed on different days
``the effect is unrecoverable''; and \texttt{pysam} fixes that ``Numeric
coordinates accepted by pysam APIs are 0-based, half-open''.
We wrote most of them ourselves (Appendix~\ref{app:anatomy}), so this is a
description of our own artifact rather than an independent audit.
Sections~\ref{sec:what} and~\ref{sec:coverage} describe what a skill is and
what the library contains, so readers can judge whether it is relevant to
their work.
We then measure two properties of the documentation files:
how well the short descriptions that remain in context are separated lexically
(Section~\ref{sec:routing}), and whether the skills named by the library's
example workflows fit in a context window (Section~\ref{sec:budget}).
We recompute every quantity from the repository tree at a pinned release
instead of taking it from our own documentation.
Section~\ref{sec:limitations} states what has and has not been measured.
The appendices contain the full descriptive census, the measurement
conventions, and the validation and scope ledgers.

This is a resource paper: its purpose is to introduce the library, not to
validate the effect of installing any individual skill.
We have published our own benchmarks of several skills at
\url{https://www.k-dense.ai/benchmarks}, but those studies do not validate the
collection.
Each skill needs evaluation on tasks and outcomes appropriate to its domain,
and broader independent work by the communities represented here remains
necessary.

\subsection{Related work}

Recent systems write and run scientific code for autonomous
chemistry~\citep{boiko2023autonomous}, hierarchical multi-agent bioinformatics
analysis~\citep{li2025kdenseanalyst}, and end-to-end research
automation~\citep{lu2024ai}; others package biomedical research workflows over
a shared tool layer~\citep{tooluniverse2026}.
Coding agents are also evaluated on repository-scale
tasks~\citep{jimenez2024swebench} and scientific
programming~\citep{tian2024scicode}.
Error analysis of a data-driven discovery benchmark finds that most failures
are programs that execute but are semantically incorrect, often because they
misuse discipline-specific tools, rather than programs that fail to
run~\citep{chen2024scienceagentbench}.

Prior approaches put knowledge into a model's context through corpus
retrieval~\citep{lewis2020rag}, tool
interfaces~\citep{schick2023toolformer,yao2023react}, or routines accumulated
by the agent itself~\citep{wang2023voyager}.
Skills differ from retrieval because each skill is an authored procedure rather
than a passage.
They differ from tools because a skill provides no runtime to call.
Retrieval can degrade in long contexts~\citep{liu2024lost}, so the tiered design
in Section~\ref{sec:what} affects how the library operates, not just how it is
presented.

Recent work examines how skill libraries are built, catalogued and
evaluated~\citep{shen2026skillfoundry,li2026skillsbench,ling2026datadriven},
how they behave as they grow~\citep{li2026dynamic}, and what happens when an
agent must retrieve its own skills from a large uncurated
collection~\citep{liu2026realistic}, and what happens when its tool layer
already returns strict, schema-validated
observations~\citep{chacko2026negative}.
The retrieval study and the re-analysis both report benefits shrinking toward
the no-skill score, the former once the agent must select the skill itself.
Section~\ref{sec:routing} therefore measures how distinct the short
descriptions that stay in context are.
Other work treats agent-loadable instruction files as an attack
surface~\citep{greshake2023injection,jiang2026sok,liu2026wild,li2026secure,
cisco2026skillscanner,holzbauer2026context}.
Appendix~\ref{app:validation} places the library's security material in this
context.
Appendix~\ref{app:method} relates skills to workflow engines and data
standards, which address the same procedural problem for pipelines.

\section{What an agent skill is}
\label{sec:what}

A skill is a directory.
At minimum it contains \texttt{SKILL.md}: a YAML header followed by Markdown
instructions written for an agent rather than a human reader.
The header is constrained to six fields: \texttt{name},
\texttt{description}, \texttt{license}, \texttt{compatibility},
\texttt{allowed-tools} and \texttt{metadata}.
Any other top-level key is a specification error.
The specification recommends three conventional subdirectories:
\texttt{references/} for additional documentation, \texttt{scripts/} for code,
and \texttt{assets/} for templates and static resources, while permitting
other files and directories.
We adopt a stricter layout and allow only \texttt{SKILL.md} and those three
subdirectories inside each skill (Appendix~\ref{app:design}).
Appendix~\ref{app:anatomy} describes the directory structure and
instruction-file contents, and shows a skill in its shipped form.

For the skills we write ourselves, our authoring workflow usually begins with a
language-model draft.
We then review the text manually and test the documented workflow for basic
functionality when that workflow's environment is available.
We revise each skill as we learn when it works and when it does not, drawing
on our own use, community feedback and outside contributions.
These steps are authoring and maintenance, not domain validation;
Appendix~\ref{app:validation} states what the automated checks cover.

\subsection{Progressive disclosure}

\begin{figure}[htbp]
  \centering
  \includegraphics[width=\textwidth]{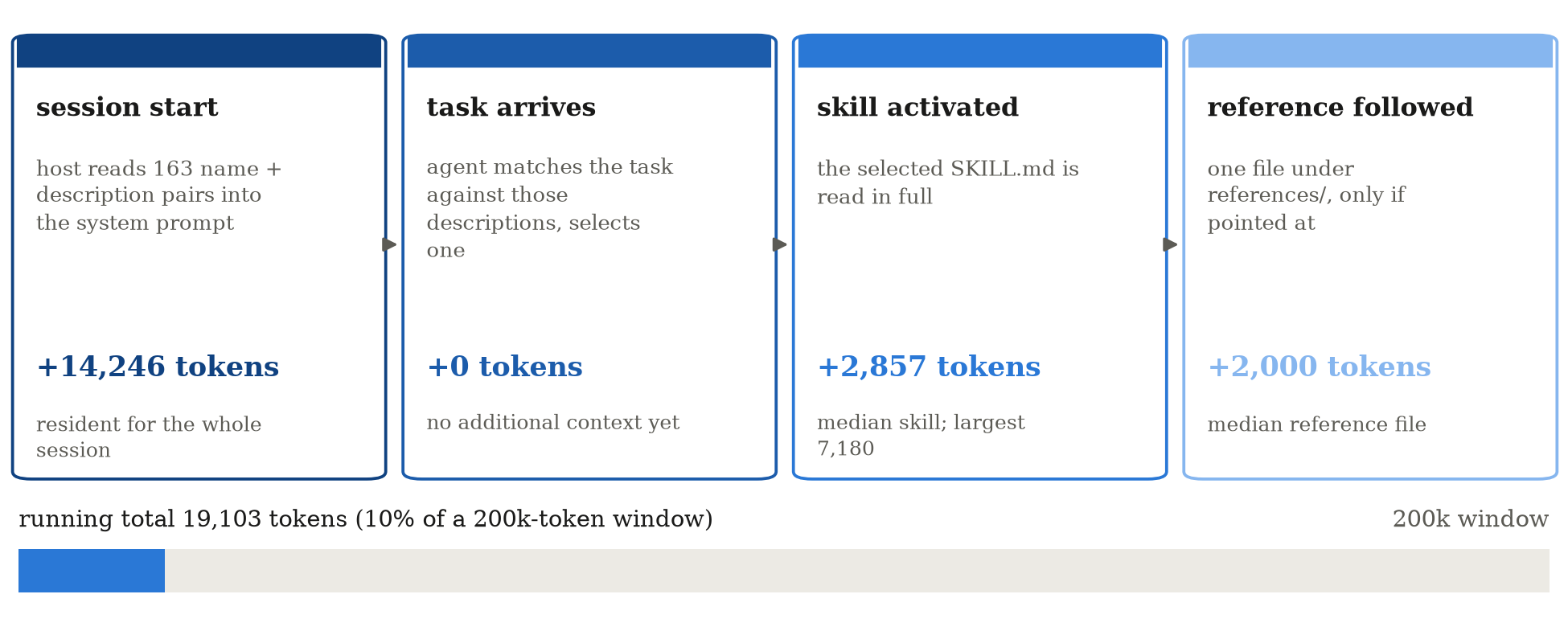}
  \caption{Measured token costs of the three disclosure tiers at tag \PinTag{},
  shown along a hypothetical task path. These are corpus measurements, not an
  agent trace. Selecting a skill uses descriptions already in standing context,
  so it adds no documentation tokens; Section~\ref{sec:routing} treats that
  step separately. The running total assumes a median skill and one median
  reference file. Section~\ref{sec:budget} costs the library's documented
  workflows instead.}
  \label{fig:trace}
\end{figure}

The design keeps almost all library content out of standing context.
Under the specification's disclosure model, an agent host reads the
\texttt{name} and \texttt{description} of every installed skill into its system
prompt at startup.
What a host adds around those entries is host-specific, as discussed in
Section~\ref{sec:limitations}.
The instruction body is read when the agent judges the skill relevant; a
reference document is read only if the instruction body points at it.
The disclosure model has three tiers whose sizes differ by orders of magnitude
(Figure~\ref{fig:trace}).

Across the whole library, the resident tier is \TierOneTokens{} tokens, a
median of \TierOneMedianTokens{} tokens per skill.
The complete instruction files behind the resident tier total
\TierTwoTokens{} tokens, with a median of \TierTwoMedianTokens{} tokens each.
The reference documents behind \emph{those} instruction files total
\TierThreeTokens{} tokens across \NumReferenceFiles{} files.
As a result, \FracDeferred\% of the library's documentation remains unread
unless an activated skill points to it.
Holding the entire library available costs \TierOneAsFracCorpus\% of the
\DocTokensTotal{}-token corpus in standing context, or
\ResidentFracWindow\% of a \WindowTokens{}-token reference window.
This low resident cost allows a host to offer \NumSkills{} specialised
procedures without filling its context window with procedures the current task
does not need.

Tiering defers cost; it does not remove it.
A tier the agent never reads cannot inform its choice, so routing for the whole
library relies on the resident tier alone.
Reference material saves context only while it remains deferred during a real
task.
Sections~\ref{sec:routing} and~\ref{sec:budget} measure these constraints.
Both measurements describe the documentation layout rather than agent traces:
they show what selection costs and where it may be difficult, not whether any
agent selects correctly.
Appendix~\ref{app:method} states the tokenizer and the corpus definition the
figures depend on.

\section{What the library covers}
\label{sec:coverage}

\begin{figure}[htbp]
  \centering
  \includegraphics[width=\textwidth]{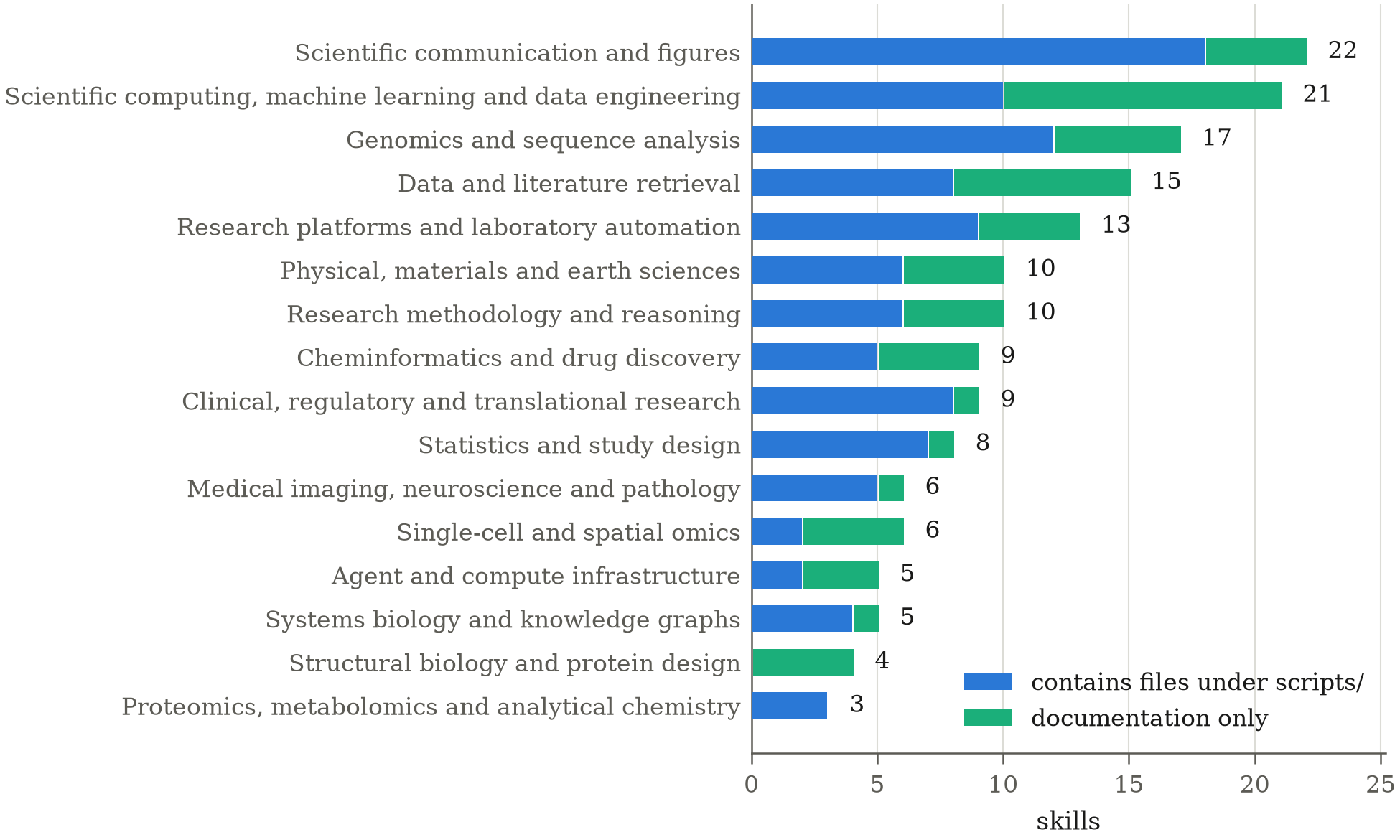}
  \caption{Skills per category, split by whether the skill ships files under
  \texttt{scripts/}. Every such skill has a corresponding test-suite directory.
  The existence of a suite does not establish that every helper or code path is
  exercised; a documentation-only skill leaves implementation to the agent.}
  \label{fig:overview}
\end{figure}

The library is organised as one directory per skill under \texttt{skills/},
with no enforced grouping above that level.
Our own documentation groups skills in three different, partly overlapping
ways, none of which partitions the library.
For reporting, we assign each skill exactly one primary category.
This taxonomy is our reporting convention, not a property of the repository.
Appendix~\ref{app:coverage} gives the full taxonomy and per-category counts.
At this tag the \NumSkills{} skills span \NumCategories{} categories, the
largest being scientific communication and figures, scientific computing and
data engineering, and genomics and sequence analysis.
Of these, \NumWithScripts{} ship executable code under \texttt{scripts/} and
\NumWithReferences{} ship reference documents (Figure~\ref{fig:overview}).

\begin{figure}[htbp]
  \centering
  \includegraphics[width=\textwidth]{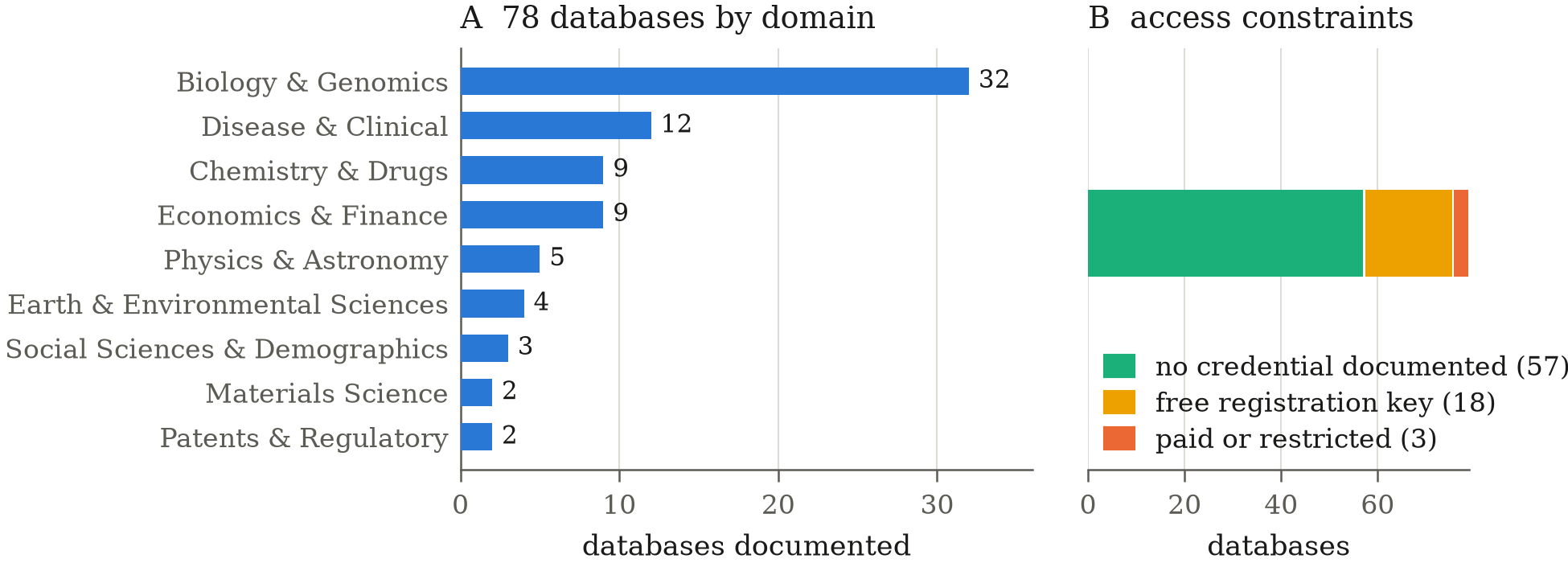}
  \caption{The \NumDatabasesDocumented{} databases documented by the
  \texttt{database-lookup} skill at tag \PinTag{}. \textbf{(A)} Databases by
  domain, using the skill's own grouping, which we checked against the
  reference files on disk.
  \textbf{(B)} Access constraints from the skill's own tables:
  \DatabasesOpen{} databases document no credential requirement,
  \DatabasesKeyed{} require a free-registration key, and
  \DatabasesRestricted{} are paid or otherwise restricted; each restricted
  database has a named free alternative.}
  \label{fig:databases}
\end{figure}

The library includes four recurring skill types across categories.
\emph{Package workflows} document how to use a specific scientific package
correctly, such as Scanpy~\citep{wolf2018scanpy} for single-cell analysis or
PyDESeq2~\citep{muzellec2023pydeseq2} for differential expression; their
content is closest to authoritative documentation.
\emph{Data retrieval} skills handle identifier resolution and query construction
against named resources; one skill, \texttt{database-lookup}, alone documents
\NumDatabasesDocumented{} databases across \NumDatabaseDomains{} domains, of
which \DatabasesOpen{} document no credential requirement and
\DatabasesKeyed{} require a key (Figure~\ref{fig:databases}).
\emph{Research platforms and laboratory automation} skills address instrument
and platform interfaces.
They are the group most likely to require credentials: across the library
\NumCredentialDeclaring{} skills name one of \NumCredentialVars{}
credential-like environment variables.
\emph{Method and judgment} skills encode procedure rather than tooling, such as
study design and randomisation before data exist, sample-size
justification~\citep{button2013power}, test selection, assumption checking and
the interpretation of significance thresholds~\citep{wasserstein2016asa}, and
uncertainty and unit propagation.
The \texttt{experimental-design} skill, for instance, lists ``mistakes that
ruin studies'' which ``can't be fixed in analysis, only in design'', beginning
with pseudoreplication (``The replicate must be at the level the treatment is
randomized'') and nuisance-variable confounding (``Randomize across, or block
on, every nuisance factor you can name'').
This group is the hardest to evaluate because its value is least like
documentation lookup.
Appendix~\ref{app:coverage} describes all four in more detail.

\begin{figure}[htbp]
  \centering
  \includegraphics[width=\textwidth]{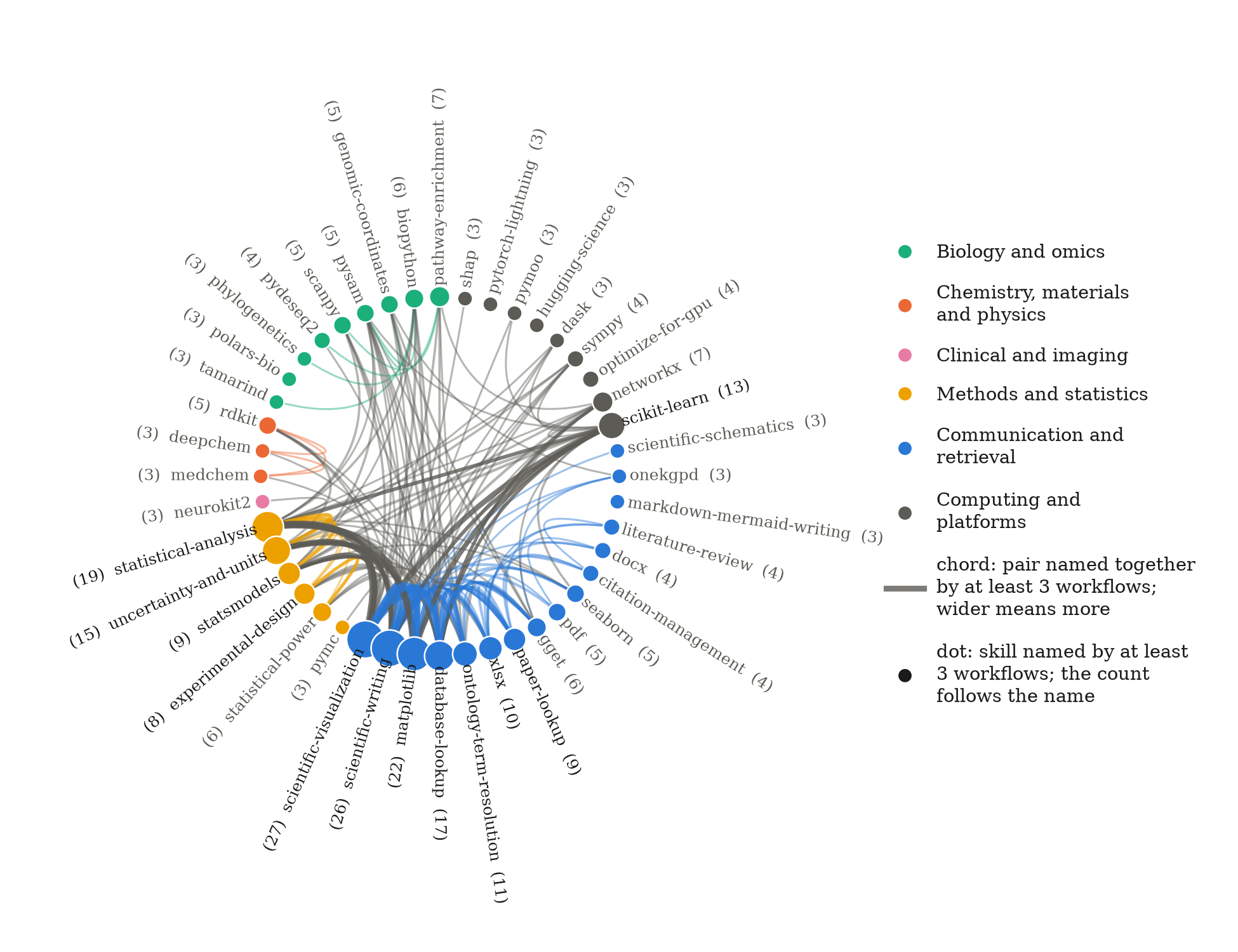}
  \caption{How the \NumDocWorkflows{} documented workflows compose skills at
  tag \PinTag{}. The \CompositionCoreSkills{} skills named by at least
  \CompositionMinWorkflows{} workflows are arranged by domain group, with the
  number of workflows naming each skill after its name. A chord joins each of
  the \CompositionChords{} pairs that at least \CompositionMinShared{}
  workflows name together, and its width grows with that count. The six groups
  colour our \NumCategories{}-category taxonomy and are used only here. The
  figure describes our own worked examples, not observed agent sessions.}
  \label{fig:workflows}
\end{figure}

The documented workflows show how these skills combine.
Each of the \NumDocWorkflows{} worked examples in \texttt{docs/examples.md}
names the skills it draws on, and the median example spans
\WorkflowCategoriesMedian{} of the \NumCategories{} categories
(Figure~\ref{fig:workflows}).
Composition is uneven: \WorkflowSkillsUsedOnce{} of the \WorkflowSkillsUsed{}
skills named appear in exactly one example, whereas
\texttt{\MostComposedSkill{}} appears in \MostComposedSkillWorkflows{} and
\texttt{\SecondComposedSkill{}} in \SecondComposedSkillWorkflows{}.
In these examples the communication and statistics skills are the connective
tissue and the domain skills are what they connect.

\section{Can descriptions distinguish among skills?}
\label{sec:routing}

Progressive disclosure keeps standing context small only if a
\TierOneMedianTokens{}-token resident description can distinguish one skill
from \NumSkills{} others.
The library does not enforce this requirement, and no host reports a selection
rate.
We measure the lexical distance between descriptions and identify pairs for
which we added explicit guidance.

We represent each description as a bag of lower-cased alphanumeric tokens that
are at least \RouteMinTokenLen{} characters long.
We remove a \RouteStopWords{}-word function-word list, weight the remaining
terms by log frequency times inverse document frequency, and calculate cosine
similarity for all \RoutePairsTotal{} pairs.
This is a lexical proxy for confusability.
It locates pairs built from the same words; it does not establish that any host
would confuse them, as Section~\ref{sec:limitations} makes explicit.
Because the preprocessing is our convention, Appendix~\ref{app:sensitivity}
reports how far the ranking moves under three alternatives, including no stop
list.

\begin{figure}[htbp]
  \centering
  \includegraphics[width=\textwidth]{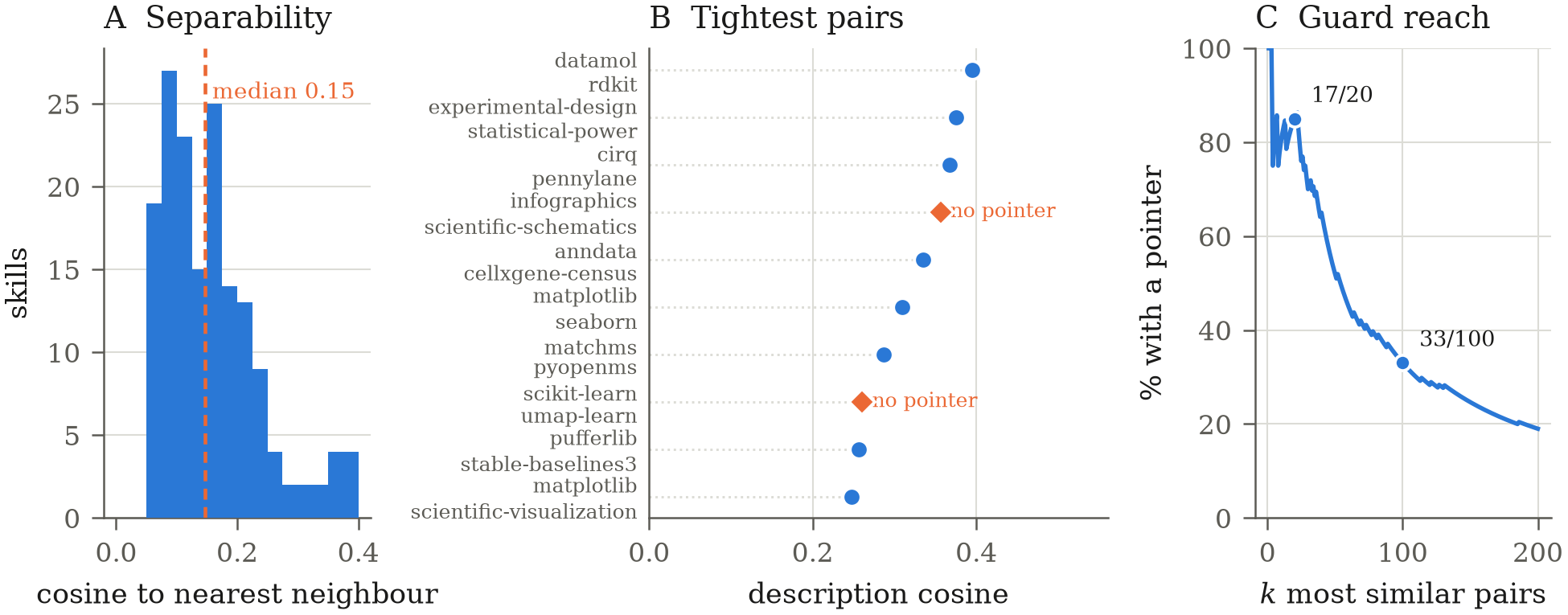}
  \caption{Lexical separability of the always-resident descriptions at tag
  \PinTag{}. \textbf{(A)} Each skill's cosine to its nearest neighbour; most
  descriptions have no lexically close competitor. \textbf{(B)} The ten
  closest pairs, marked by whether either description points to the other.
  \textbf{(C)} The share of the $k$ most similar pairs carrying such a pointer.
  Cosine is a lexical proxy for confusability, not a measurement of any host's
  selection behaviour.}
  \label{fig:routing}
\end{figure}

Most descriptions are lexically well separated.
The median skill's nearest neighbour has a cosine similarity of
\RouteMedianNN{}, and the ninetieth percentile is \RoutePNinetyNN{}
(Figure~\ref{fig:routing}A), so a typical skill has no lexically close
competitor.
The difficult cases lie in the upper tail.
The closest pair in the library, \texttt{\RouteClosestA{}} and
\texttt{\RouteClosestB{}}, reaches \RouteClosestCos{}.
The next most similar pairs also connect adjacent work in the same domain: two
cheminformatics toolkits, two study-design skills, two quantum-computing
frameworks and two plotting libraries (Figure~\ref{fig:routing}B).
Their lexical overlap reflects adjacent work, not authoring errors.
In such cases, a \TierOneMedianTokens{}-token description must distinguish the
skills on its own.

We add cross-references to some similar descriptions.
In all, \RoutePointerSources{} descriptions name another skill directly.
Together, they contain \RoutePointerEdges{} pointers connecting
\RoutePointerPairs{} distinct pairs, of which \RoutePointerReciprocal{} are
reciprocal.
These pointers are routing hints in the resident tier: they tell the agent when
to prefer a neighbouring skill.
They are concentrated among similar descriptions.
The mean cosine of a pair joined by a pointer is \RouteMeanCosPointer{},
compared with \RouteMeanCosNoPointer{} for a pair without one, an enrichment of
about \RouteEnrichment{} times.
Pointers cover \RouteGuardedTopTwenty{} of the twenty closest pairs,
\RouteGuardedTopFifty{} of the fifty closest, and
\RouteGuardedTopHundred{} of the hundred closest
(Figure~\ref{fig:routing}C).
We found many of the closest collisions ourselves, but fewer pairs carry a
pointer farther down the ranking.

Some highly similar pairs have no guard.
At \RouteUnguardedCos{}, \texttt{\RouteUnguardedA{}} and
\texttt{\RouteUnguardedB{}} are the closest unguarded pair.
They rank as high as the closest guarded pairs, not down in the tail, yet
neither description mentions the other.
Because each guard is authored per skill rather than derived from the corpus,
coverage depends on whether the description author notices the collision.
Running this measurement on each release would identify such pairs
systematically.

Our continuous integration also computes a different but related signal.
The third-party security scanner we run has a rule that reports skills with
complementary descriptions, and it flags \RouteScannerPairs{} distinct pairs.
That rule keys on shared \emph{function} words, so the pairs it flags have a
mean cosine of \RouteScannerMeanCos{}, compared with
\RouteScannerUnflaggedCos{} for unflagged pairs.
It catches \RouteScannerTopTwenty{} of the twenty closest pairs in the library.
The two measurements are not substitutes.
Appendix~\ref{app:validation} treats the scanner's output as the heuristic it
is.

\section{Do the skills a workflow needs fit in context?}
\label{sec:budget}

The corpus reveals selection \emph{cost}, even though it cannot reveal
selection accuracy.
The \NumDocWorkflows{} worked examples in \texttt{docs/examples.md}
(Section~\ref{sec:coverage}) together contain \WorkflowSlots{} skill slots
covering \WorkflowSkillsUsed{} of the \NumSkills{} skills.
They are our only statement of how many skills a realistic task is expected to
need.
The same document opens with prompting guidance whose first rule is ``Name the
skills you want'', because a host selects skills by matching the request
against each description, the step Section~\ref{sec:routing} measures.
The median example names \WorkflowMedianSkills{} skills; the smallest names
\WorkflowMinSkills{} and the largest names \WorkflowMaxSkills{}.
We treat each example as a session to translate the per-skill token counts in
Section~\ref{sec:what} into context-window costs.

\begin{figure}[htbp]
  \centering
  \includegraphics[width=\textwidth]{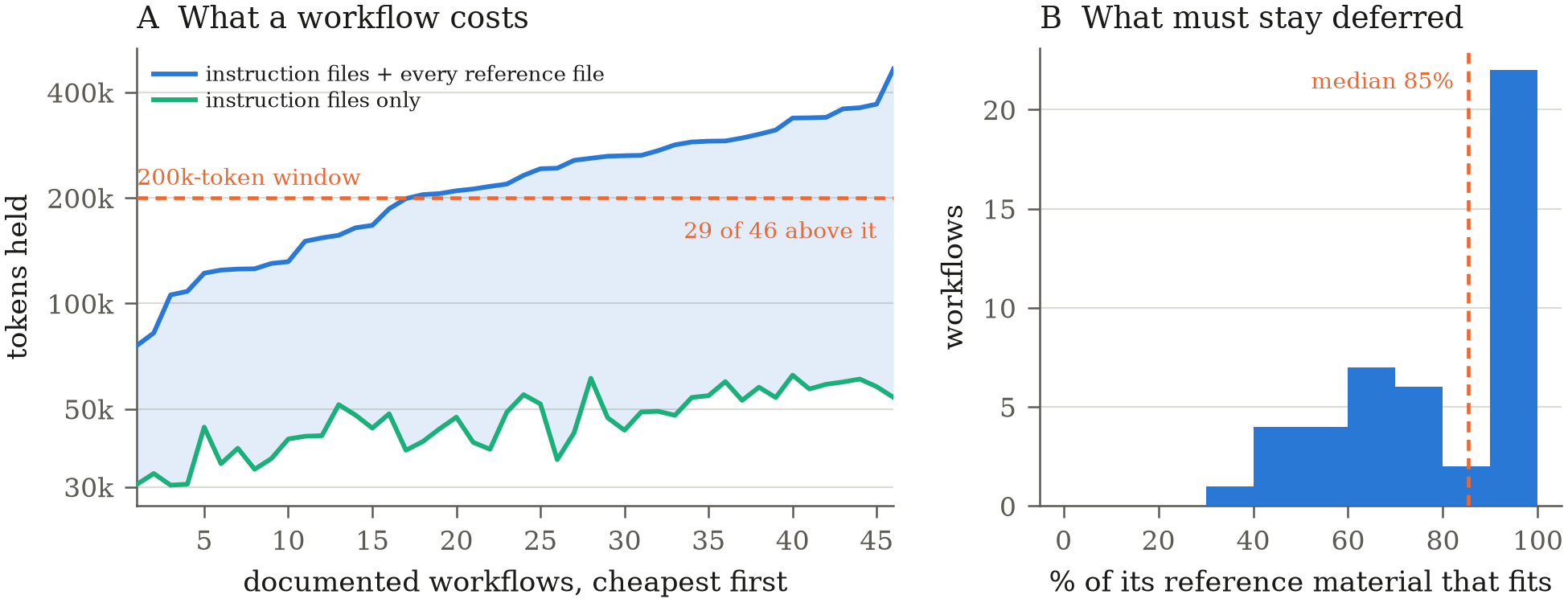}
  \caption{Token costs for the library's \NumDocWorkflows{} documented
  workflows at tag \PinTag{}, compared with a \WindowTokens{}-token reference
  window. \textbf{(A)} Cost with the resident tier and instruction files, and
  after adding every reference file of every named skill; the shaded gap is the
  deferred tier. The upper series is an upper bound by construction.
  \textbf{(B)} The share of its own reference material each workflow could load
  before exhausting the window. Both panels count documentation text only,
  excluding scripts, assets, tool output and host wrappers.}
  \label{fig:budget}
\end{figure}

The instruction files fit comfortably.
After including the resident tier, the median documented workflow costs
\WorkflowMedianInstruction{} tokens, or \WorkflowMedianInstructionPct\% of a
\WindowTokens{}-token window, and the largest costs
\WorkflowMaxInstruction{}.
\WorkflowOverWindowInstruction{} of the \NumDocWorkflows{} workflows exceed
the window (Figure~\ref{fig:budget}A).
Loading every reference file of every named skill changes the result.
The median rises to \WorkflowMedianFull{} tokens, or
\WorkflowMedianFullPct\% of the same window, and the largest rises to
\WorkflowMaxFull{}.
\WorkflowOverWindow{} of the \NumDocWorkflows{} workflows then exceed the
window.

The gap between the two series measures how much context progressive disclosure
can save on these documented workflows rather than on a hypothetical workload.
The upper series loads material the design is meant to leave on disk, so it is
an upper bound by construction.
For \WorkflowOverWindow{} of the \NumDocWorkflows{} workflows, the upper series
exceeds the \WindowTokens{}-token window, so those workflows still need
references to be loaded selectively.

The remaining context window sets a budget for reference material.
After loading a workflow's descriptions and instruction files, the remaining
tokens are available for references.
The median documented workflow can afford \WorkflowAffordMedian\% of its own
reference material, the lower quartile can afford \WorkflowAffordQOne\%, and
the most demanding workflow can afford \WorkflowAffordMin\%
(Figure~\ref{fig:budget}B).
Of the \NumDocWorkflows{}, \WorkflowAffordAll{} can afford all their reference
material; the others require selective loading, including
\WorkflowAffordUnderHalf{} that can afford under half.

These figures count documentation text only and exclude scripts, assets, tool
output and whatever framing a host wraps around each entry, whose context
exposure depends on the host and task, so they understate the amount of
context a session would hold.
Each workflow total includes every resident description plus the complete
instruction files, so an activated skill's own description is counted twice.
That extra copy is a median of \TierOneMedianTokens{} tokens; we retain it
because it biases the total upward.

We recommend installing only the skills relevant to a project rather than the
full library at once.
Every installed skill adds an always-resident description and another routing
candidate even if it is never used; limiting the installed set bounds both
standing context and the number of candidates the host must distinguish.
Whether a smaller installation improves selection accuracy remains untested.

\section{Limitations}
\label{sec:limitations}

\paragraph{No task-level evaluation.}
We describe the library, its automated checks and two corpus properties, but do
not show that giving an agent these skills improves its scientific work.
A controlled study would hold the agent, model, tools and task fixed, vary only
the installed skills, and score outcomes against answer keys based on published
conventions rather than on the skill text.
Matched evaluations with and without skills now exist for other
collections~\citep{li2026skillsbench,shen2026skillfoundry,shaposhnikov2026evaluation}.
A re-analysis of a controlled offensive-cybersecurity study finds the benefit
collapsing in that domain~\citep{chacko2026negative}.
Its authors propose that strict, schema-validated tool observations can supply
the procedural correction that a skill would otherwise provide.
Such studies also risk maintainer selection bias in their tasks, which
statistical care cannot remove.
If a capable model already completes the tasks without skills, there is no
headroom left to show a benefit, and skipping that check can yield a confident
null that reflects the task set rather than the skills.
Our own matched evaluation of a different kind of skill file, an expert
profile compiled from a person's public record, left judgment transfer
unresolved for exactly this reason: every condition, with or without the file,
scored near the ceiling on both tests~\citep{kassis2026mimeo}.

\paragraph{Selection is measured on the corpus, not in an agent.}
Section~\ref{sec:routing} maps lexical confusability and authored guards, not
how often a host selects correctly.
That rate would require labelled task-to-skill pairs, a fixed host and a scored
selection step.
Lexical cosine is computable and auditable, but remains a proxy: descriptions
can share vocabulary without confusing a model, or share none while competing
for the same task.
It maps lexical similarity but does not estimate an error rate.
One study finds that skill benefits can shrink toward the no-skill score when
an agent retrieves skills from a large uncurated
collection~\citep{liu2026realistic}; another reports that flat retrieval can
degrade as libraries grow~\citep{li2026dynamic}.
Their different library sizes and curation methods do not predict a rate here,
but show why one must be measured.

\paragraph{A selected skill can still fail scientifically.}
Selection does not show that an agent followed the skill: it may omit or
reorder steps, skip a stop condition, or load only part of the references.
Even faithful execution can be wrong when the study violates the procedure's
assumptions.
The task or data may omit relevant details about study design, sampling,
measurement, preprocessing, population or estimand, and a skill cannot apply
conditions the host does not reveal.
Skills and their dependencies can also become stale as standards, software,
APIs, database schemas and reference data change.
A pin records the available instructions, not their correctness or currency.
Reproducing a run also requires the relevant data, software and environment
details, plus database access dates where applicable.
We measure neither step-level adherence nor contextual sufficiency, and do not
audit every procedure or dependency against current domain guidance.
Skills structure scientific judgment; they do not replace expert review.

\paragraph{Reach is not use.}
The installed-base estimate in Appendix~\ref{app:base} is the only number not
derived from the pinned tree, and it counts clients that fetch the library, not
scientists or sessions in which an agent used a skill.
It is not evidence of benefit, and nothing else in the paper rests on it.

\paragraph{Host portability is intended, not tested.}
Conformance to a shared layout does not show that hosts discover, select or
execute a skill identically: wrappers, install locations, metadata, tool names
and permission models differ.
A compatibility claim would require a host-by-skill matrix that this paper does
not provide.

\paragraph{Uneven review and incomplete CI coverage.}
Across the library, \NumExternallyAuthored{} of \NumSkills{} skills do not list
us as their author, including \NumUnattributed{} with no named author; the
latter are unattributed, not known to be external.
Our own security guidance warns that community contributions may receive less
review than the skills we write, so the corpus is not uniformly vetted.
CI also leaves conventions unenforced: the structural contract covers only
\NumStructurallyChecked{} skills, leaving \NumStructuralFindings{} broken local
paths outside its gate (Appendix~\ref{app:validation}).

\paragraph{Our measurements depend on stated conventions.}
Our choices determine the \NumCategories{}-category taxonomy, the lexical
scope-candidate rule and adjudication, the heading-role regexes, the
credential-name rule, the routing preprocessing, the workflow parse and the
two thresholds of the composition figure.
Only the adjudicated scope positives are a lower bound.
Heading roles and credential variables are lexical counts; the other items are
analysis conventions, and every token figure depends on the fixed tokenizer.
We describe each rule rather than publishing its word lists, so these
measurements can be approximated from the tagged tree but not reproduced
exactly; Appendix~\ref{app:sensitivity} varies the routing rules and reports
the effect.

\paragraph{A fixed snapshot of a changing library.}
The most recent \ReleasesInWindow{} releases returned by the API span
\ReleaseWindowDays{} days, about \ReleasesPerMonth{} releases per month, but
that page is a recent window rather than the complete history.
Every count describes tag \PinTag{} and will differ from later releases, so we
state the pin and regenerate each number from the tree.

\section{Availability and citation}

Scientific Agent Skills is available at
\url{https://github.com/K-Dense-AI/scientific-agent-skills} under the MIT
licence.
This paper describes tag \PinTag{}, released \PinDate{}, at commit
\texttt{\PinSha{}}.
At the pin, our README names Claude Code, Claude Cowork, Codex, Gemini CLI,
Google Antigravity and Cursor as supported hosts, a claim
Section~\ref{sec:limitations} qualifies, and recommends installing with
\texttt{npx skills add K-Dense-AI/scientific-agent-skills}; a release can be
pinned at install time with \texttt{gh skill install
K-Dense-AI/scientific-agent-skills -{}-pin \PinTag{}}, which records the tag a
citation should name.
At the time of writing (\StarsDate{}) the repository has \Stars{} GitHub stars;
Appendix~\ref{app:base} reports a \TrafficWindowDays{}-day traffic window and
estimates roughly \BaseEstimateRounded{} distinct cloning clients behind it, a
proxy for reach that counts neither people nor agent sessions.

Work that uses these skills should cite this paper together with the release
tag it used.
The library's contents change between releases, so a citation without a tag
does not identify what was run.
A paper reporting a result obtained with a specific skill should also name that
skill and its \texttt{metadata.version}, which is maintained independently of
the repository version.
For skills in the clinical, regulatory or imaging categories, it should also
state the scope limits in that particular skill
(Appendix~\ref{app:scope}).

We do not release an analysis package with this paper.
Every corpus-derived quantity is computed from the public repository tree at
tag \PinTag{} with NumPy~\citep{harris2020array} and
Matplotlib~\citep{hunter2007matplotlib}, so the input to every measurement is
available even though our scripts are not.
The convention behind each number is described where the number is reported,
and Appendix~\ref{app:method} gives the tokenizer and corpus definition every
token figure depends on.
Recomputation would still not validate our labels or interpretation:
Appendix~\ref{app:validation} shows that a passing tree does not put every
skill under the same gates.

\paragraph{Acknowledgements.}
The skills described here document a large body of open-source scientific
software whose authors did the harder work.
The \NumVendoredSkills{} document-format skills vendored from Anthropic's public
skills repository are used under its terms.
The frontispiece was produced by an image model from our prompt; the numbered
figures are drawn in code from the measured data.

\paragraph{Competing interests.}
Timothy Kassis, Yuhuan He, Darshil Patel and Aubrey M. Brueckner are employees
of K-Dense, Inc., which maintains Scientific Agent Skills.
Vinayak Agarwal is a former employee of K-Dense, Inc.
The authors are therefore describing an artifact developed and maintained by
their current or former employer.

\bibliographystyle{unsrtnat}
\bibliography{refs}

\appendix
% Appendix floats are numbered A.1, B.1, ... and reset per appendix; body
% figures keep plain numbers. `amsmath` is already loaded for this.
\numberwithin{figure}{section}
\numberwithin{table}{section}
\setcounter{figure}{0}
\setcounter{table}{0}
% The appendices are float-dense the way the body used to be, so the loosened
% float parameters that used to sit in the preamble belong here instead.
\setcounter{topnumber}{3}
\setcounter{totalnumber}{4}
\renewcommand{\topfraction}{0.9}
\renewcommand{\textfraction}{0.08}
\renewcommand{\floatpagefraction}{0.75}

\section{Measurement conventions and the token ledger}
\label{app:method}

\paragraph{Tokenizer and corpus.}
Every token figure in this paper counts the raw text of a file under the
\texttt{\Tokenizer{}} encoding, with no host wrapper, chat template or system
prompt around it.
The count is a property of the file, so a host using a different tokenizer
would obtain a different figure; the ratios between tiers are the stable part.
The documentation corpus is tier two plus tier three: the \TierTwoTokens{}
tokens in the \NumSkills{} complete \texttt{SKILL.md} files and the
\TierThreeTokens{} tokens in the \NumReferenceFiles{} files under
\texttt{references/}, for \DocTokensTotal{} tokens in all.
The resident tier is not a third addend.
A skill's \texttt{name} and \texttt{description} live in the header of its own
\texttt{SKILL.md}, so the \TierOneTokens{}-token resident tier is an excerpt
of tier two rather than a quantity beside it, and adding it would count every
description twice.
Under \texttt{references/} we count text files only
(\texttt{.md}, \texttt{.txt}, \texttt{.rst}, \texttt{.json}, \texttt{.yaml}
and \texttt{.yml}); scripts and binary assets are counted as files but never as
documentation tokens, which is why the reported costs are lower bounds.
Editor and operating-system artifacts (\texttt{.DS\_Store}, Python bytecode and
similar) are excluded from every file count.
The \WindowTokens{}-token window used throughout is a round reference scale
rather than a property of a particular model.

\paragraph{Reproducible practice in computational science.}
Skills address a familiar problem: known procedures are often not encoded in a
reusable form.
Workflow engines such as
Nextflow~\citep{ditommaso2017nextflow} and Snakemake~\citep{koster2012snakemake},
curated software distributions~\citep{gruening2018bioconda}, the FAIR
principles for data~\citep{wilkinson2016fair}, software citation
practice~\citep{smith2016software}, and guidance on computational
craft~\citep{wilson2017goodenough} address the same problem.
These mechanisms are complementary: workflow engines make an executable
procedure reproducible, distributions make its dependencies installable, and
data standards preserve its inputs and outputs.
A skill tells an agent how to select, sequence and check those components, while
still depending on the same environments and records.

\subsection{The full token ledger}

\begin{figure}[htbp]
  \centering
  \includegraphics[width=\textwidth]{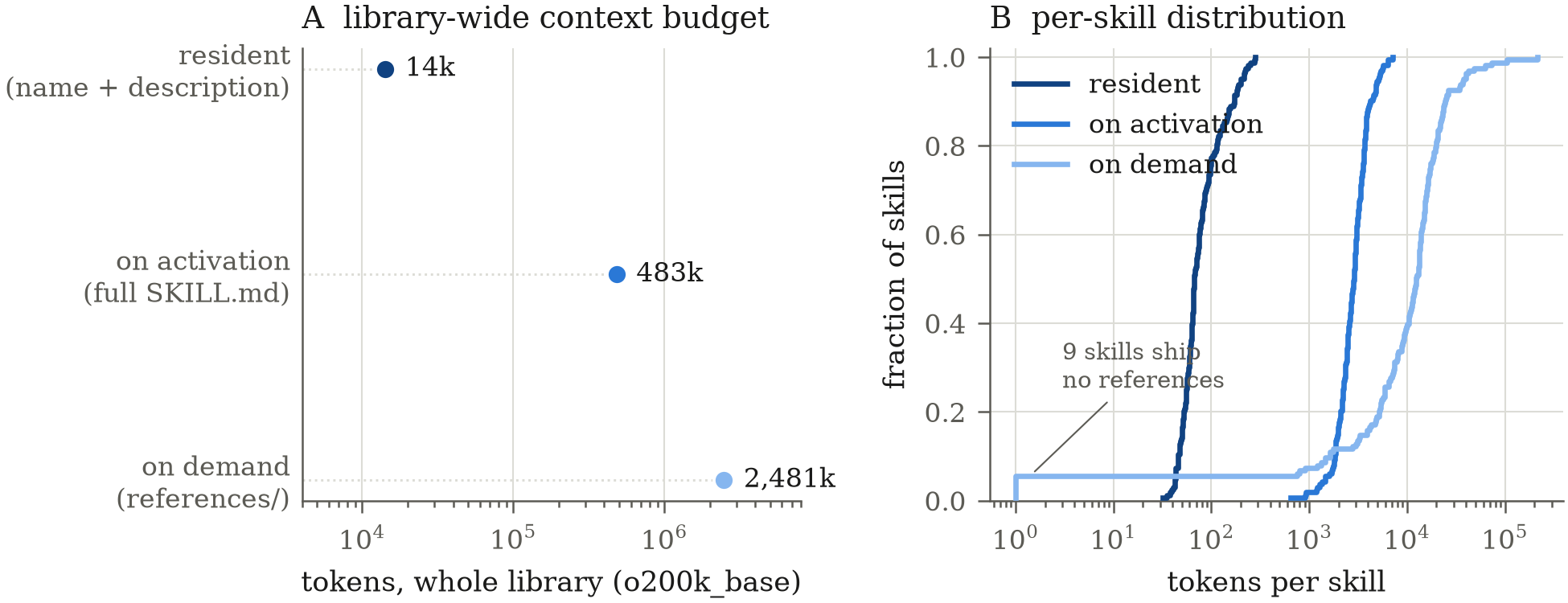}
  \caption{The three disclosure tiers at tag \PinTag{}. \textbf{(A)} What the
  whole library costs in each tier, on a log scale: the always-resident
  descriptions are \TierOneAsFracTierTwo\% of the instruction bodies, and those
  bodies are in turn a fifth of the deferred reference material.
  \textbf{(B)} The same three tiers as per-skill distributions. The deferred
  tier is heavy-tailed: a handful of skills carry most of it, which is
  why the totals alone would be misleading.}
  \label{fig:tiers}
\end{figure}

\begin{figure}[htbp]
  \centering
  \includegraphics[width=\textwidth]{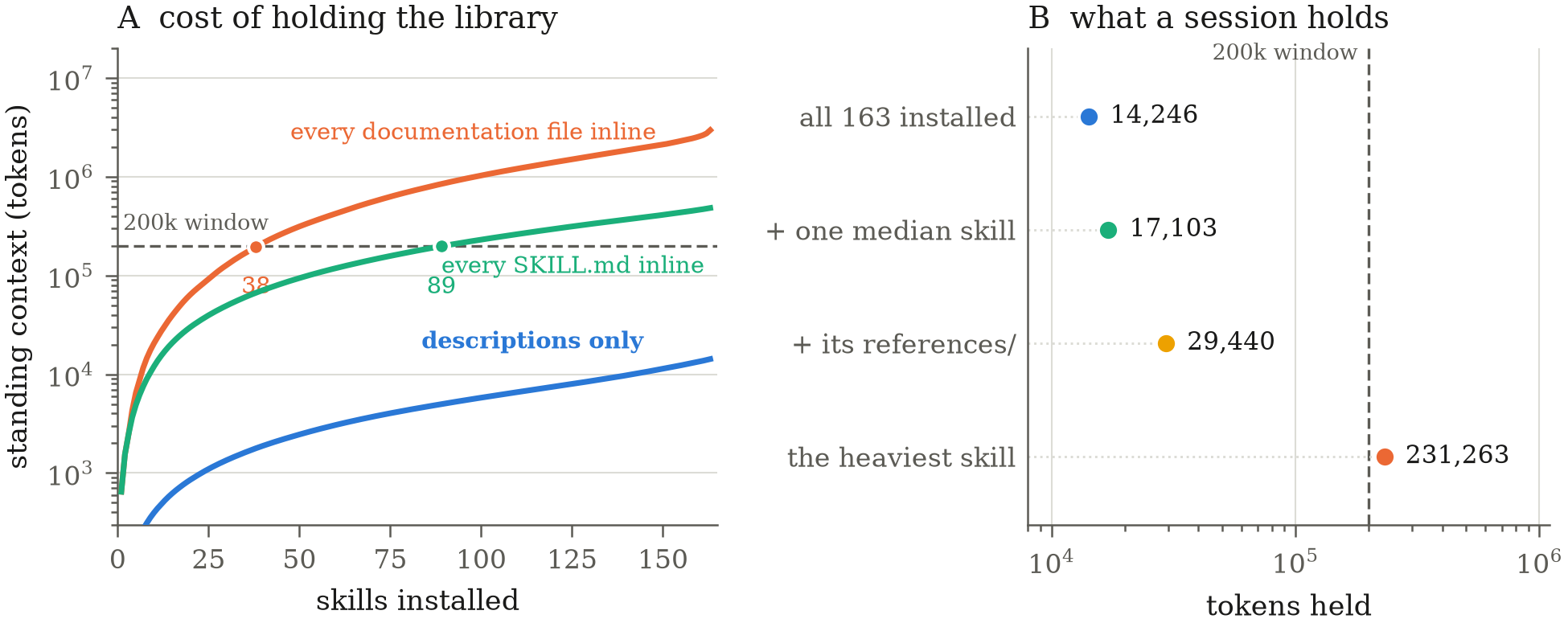}
  \caption{The cost of the library at tag \PinTag{}, against a
  \WindowTokens{}-token reference window. \textbf{(A)} Standing context as
  skills are installed, cheapest first, under three representations of the same
  material; the marked points are the last skill that fits. Cheapest-first is
  the ordering most favourable to the two inlined variants, so those marks are
  upper bounds on how many skills either could hold. \textbf{(B)} What a
  session holds in practice. Both panels use a log scale; in panel B, read a
  value from the vertical position of its dot rather than the apparent length
  of its bar.}
  \label{fig:economics}
\end{figure}

Figure~\ref{fig:tiers} shows what each tier costs, in aggregate and per skill.
Figure~\ref{fig:economics} shows the arithmetic behind tiering.
We use a \WindowTokens{}-token context window as a round reference scale, not
as a property of any particular model.
Holding the always-resident descriptions of all \NumSkills{} skills costs
\TierOneTokens{} tokens, or \ResidentFracWindow\% of that window.
Holding the same library with complete instruction files inlined would cost
\AllSkillMdResident{} tokens.
At most \FitAllSkillMd{} skills fit when the shortest files are installed
first; if each installed skill also contributes all its reference files, that
upper bound falls to \FitWholeCorpus{}.
The full documentation corpus would need \CorpusWindows{} such windows.
These counterfactuals measure documentation text only and exclude scripts,
assets, tool output and host wrappers, although those resources may still
consume context.
The full corpus therefore does not fit in one reference window.

In practice, a session holds the resident tier plus content from any skills it
activates.
With everything installed and one skill activated, the median total is
\MedianTaskTokens{} tokens, or \MedianTaskFracWindow\% of the reference window.
This total includes every description and the complete instruction file for
one skill.
Because that skill's description is already resident, the total counts it
twice, adding a median of \TierOneMedianTokens{} tokens.
We retain that double count because it biases the estimate upward.
If every activated skill were read together with its entire
\texttt{references/} directory, the median session total would instead be
\MedianTaskWithReferences{} tokens.
The heaviest skill would reach \HeaviestTaskTokens{} tokens and exceed the
window by itself.
We compute the two median session totals from their respective per-skill
distributions rather than by adding the component medians reported above.
We cost the deferred tier file by file rather than directory by directory: the
median reference file is \MedianReferenceFileTokens{} tokens
and the largest is \MaxReferenceFileTokens{}, so the choice of file can matter
more than the choice of skill.

\FloatBarrier

\section{What a skill contains}
\label{app:anatomy}

\begin{figure}[htbp]
  \centering
  \includegraphics[width=\textwidth]{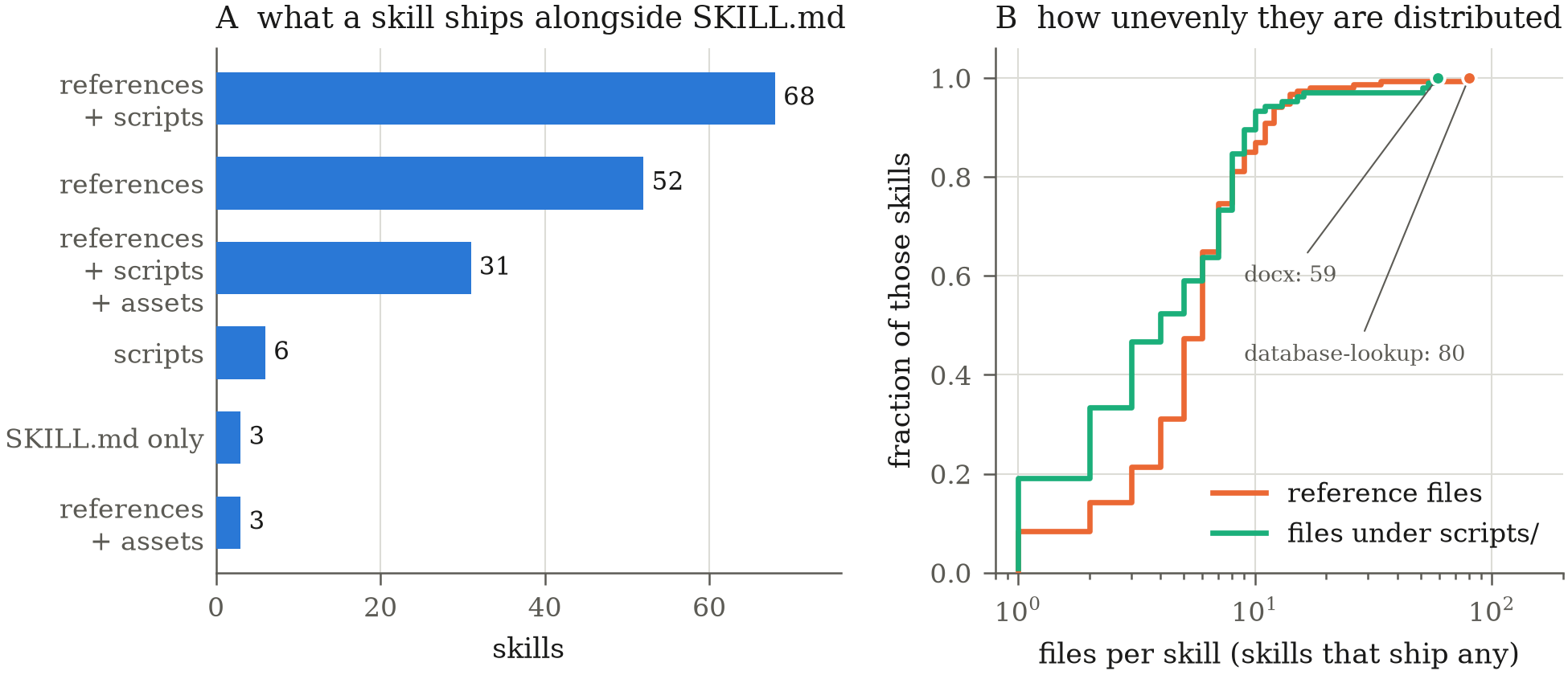}
  \caption{What skills ship at tag \PinTag{}. \textbf{(A)} The combinations of
  optional subdirectories. \textbf{(B)} Reference-file counts among skills with
  references and script-file counts among skills with scripts. Both
  distributions are heavy-tailed, so a few skills account for much of the
  library totals. The median skill with references ships
  \MedianReferencesPerSkill{} reference files; the median script-bearing skill
  ships \MedianScriptsPerSkill{} script files.}
  \label{fig:composition}
\end{figure}

Skills differ substantially in what they ship alongside the instruction file:
some bundle reference documents, scripts or assets, while others do not.
Of \NumSkills{} skills, \NumWithReferences{} ship reference documents,
\NumWithScripts{} contain \NumScriptFiles{} files under \texttt{scripts/}, and
\NumWithAssets{} ship static assets.
Those \NumScriptFiles{} comprise \NumPythonFilesInScripts{} Python files,
\NumShellFilesInScripts{} shell scripts and \NumSupportFilesInScripts{}
XML or schema support files, so the directory total is not an executable-script
count.
The most common composition is a skill with both references and scripts
(\ComboReferencesScripts{} skills); \ComboReferencesOnly{} ship references
alone, \ComboAllThree{} ship all three subdirectories, and
\ComboSkillMdOnly{} skills each consist only of a single instruction file
(Figure~\ref{fig:composition}).

A documentation-only skill explains what to do but leaves implementation to the
agent, so the output is only as reliable as the code the agent generates.
Rigorous testing shows that this generation is less reliable than a passing
example suggests~\citep{liu2023evalplus}.
A script-bearing skill can keep fragile or repetitive logic out of code
generated during a session.
Every such skill has a corresponding suite, but the existence of that directory
does not by itself establish coverage of every helper or code path.
Bundled scripts also add risk.
An ecosystem-scale survey finds that skills with executable scripts are
markedly more likely to carry a vulnerability than instruction-only
skills~\citep{liu2026wild}.
Shipped code therefore requires the testing and scanning described in
Appendix~\ref{app:validation}.
Our contributor guide identifies this as the purpose of
\texttt{scripts/} and requires every script-bearing skill to carry a test suite
(Appendix~\ref{app:validation}).

\subsection{The instruction file as a document}

\begin{figure}[htbp]
  \centering
  \includegraphics[width=\textwidth]{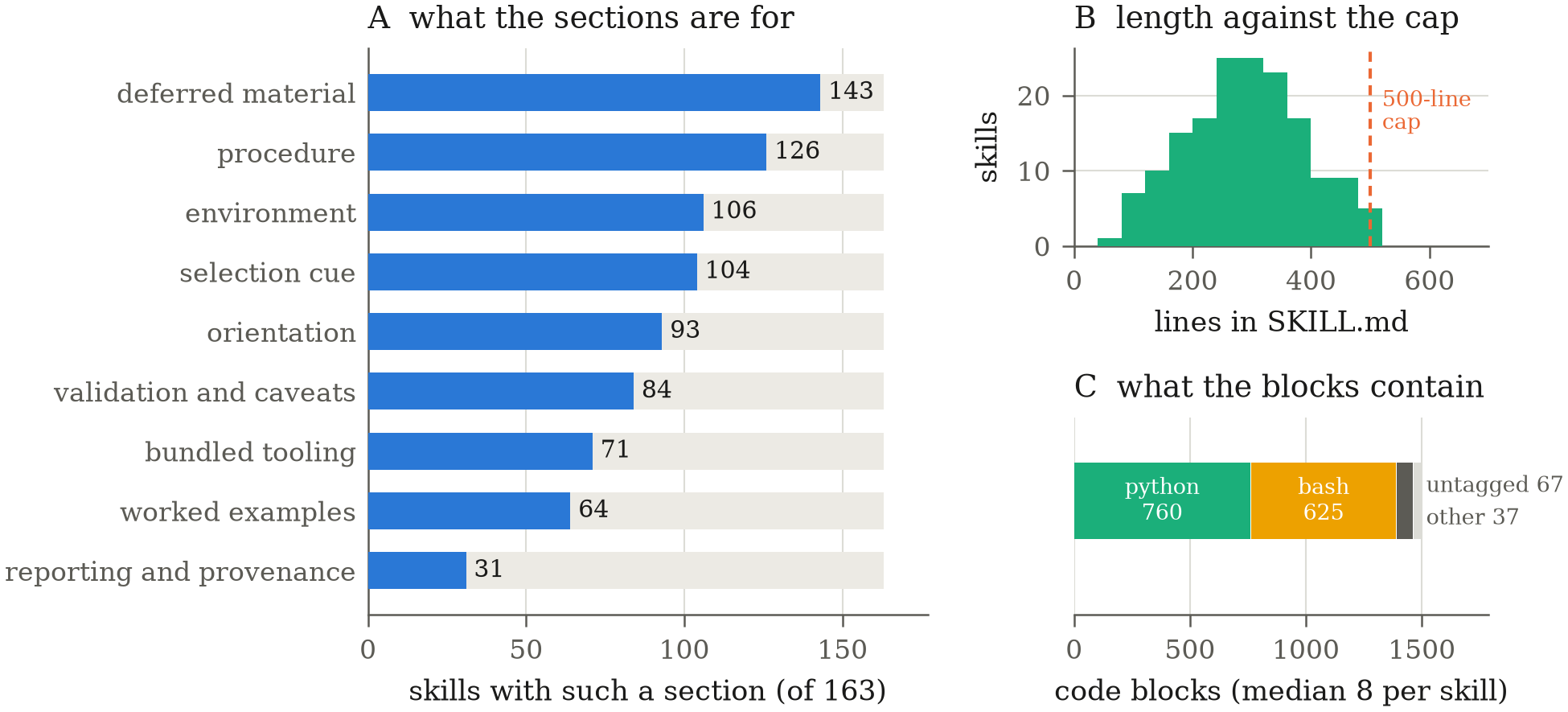}
  \caption{Inside the file an agent reads on activation, at tag \PinTag{}.
  \textbf{(A)} Level-2 sections grouped by the role they play; roles are not
  mutually exclusive. The grouping uses a keyword rule over headings rather
  than a semantic judgment. \textbf{(B)} File length against the repository's
  \SkillMdLineCap{}-line cap. \textbf{(C)} Languages declared by fenced code
  blocks; the segments account for all \TotalCodeBlocks{} blocks, including the
  \UntaggedBlocks{} that declare no language.}
  \label{fig:instruction}
\end{figure}

An agent always reads the instruction file once it selects a skill, so this file
contains the skill's main guidance.
The documents share broad structural patterns but no common template.
Across the library they use \DistinctHeadings{} distinct level-2 headings, a
median of \MedianHeadingsPerSkill{} per file, but those headings serve a small
set of recurring roles (Figure~\ref{fig:instruction}A).
\SkillsWithSelectionCue{} of \NumSkills{} skills carry a section telling the
agent when the skill applies, \SkillsWithProcedure{} give an explicit procedure
or decision rule, \SkillsWithCaveats{} carry validation checks or caveats, and
\SkillsWithDeferredSection{} point at their deferred material.
The grouping into roles is ours.
It uses a keyword rule over headings, so a skill that answers the same question
under an unmatched heading is not counted.

The median complete file is \SkillMdMedianLines{} lines against the
\SkillMdLineCap{}-line recommendation (Figure~\ref{fig:instruction}B), and
carries \MedianCodeBlocks{} fenced code blocks.
Across the library there are \TotalCodeBlocks{} such blocks, of which
\PythonBlocks{} are Python and \BashBlocks{} are shell, with
\UntaggedBlocks{} declaring no language at all
(Figure~\ref{fig:instruction}C).
These counts show how often worked commands appear, not how much of the corpus
is prose.
They are consistent with the contributing guidance's preference for concrete
workflows and commands over background explanation.

\subsection{Declared capability and provenance}

\begin{figure}[htbp]
  \centering
  \includegraphics[width=0.62\textwidth]{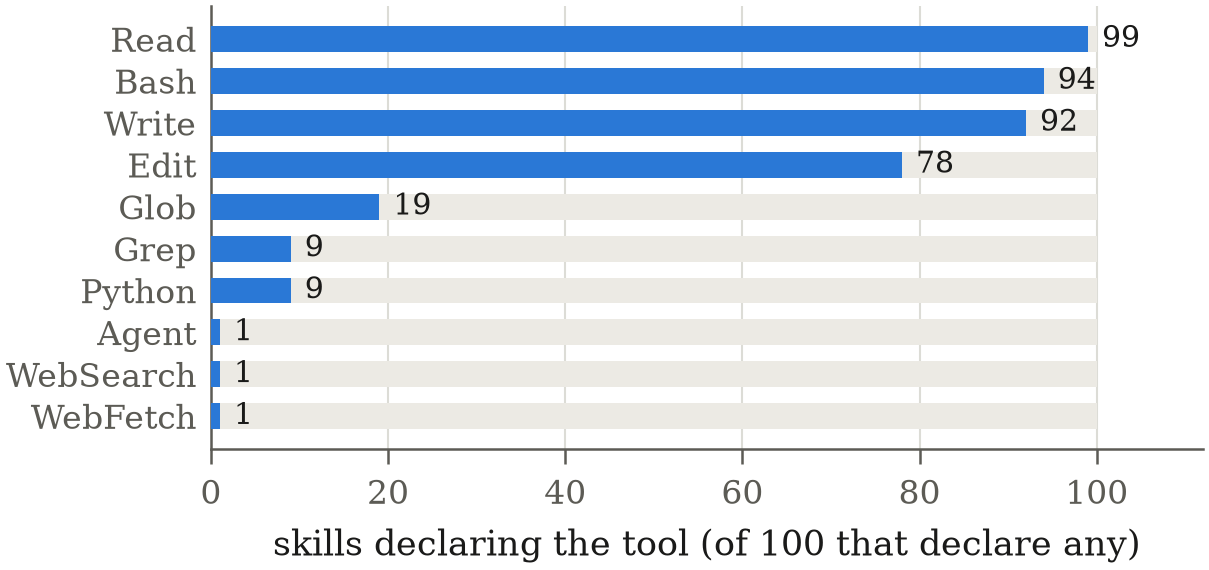}
  \caption{The host capabilities skills declare in \texttt{allowed-tools}, for
  the \NumDeclaringTools{} of \NumSkills{} skills that declare any. The field
  is optional and advisory rather than a sandbox, but where it is present it is
  the library's own statement of what a skill assumes it may do.}
  \label{fig:tools}
\end{figure}

In all, \NumDeclaringTools{} skills declare an \texttt{allowed-tools} string,
and the values are concentrated in four capabilities: \ToolRead{} ask for file
reading, \ToolBash{} for a shell, \ToolWrite{} for file writing and
\ToolEdit{} for editing (Figure~\ref{fig:tools}).
Almost all expect an agent that can read and write files and run commands.
A conversational assistant may not meet that expectation, which affects whether
the library fits a given host.
The field is optional, experimental and honoured only by hosts that implement
it.
These declarations characterise only the skills that use the field, not the
whole library.
The other \NumNotDeclaringTools{} skills state no capability requirement.
An undeclared field does not mean the skill needs no tools.

\begin{figure}[htbp]
  \centering
  \includegraphics[width=\textwidth]{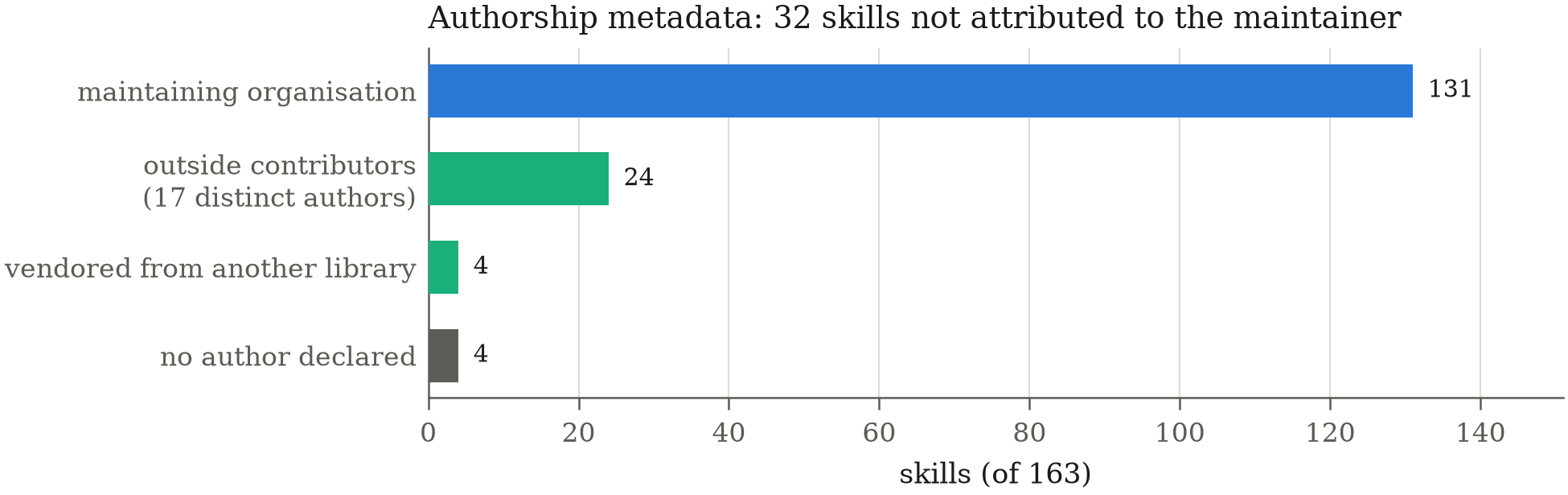}
  \caption{Authorship metadata at tag \PinTag{}, grouped rather than named.
  Per-skill attribution is in \texttt{metadata.skill-author}, which
  \NumUnattributed{} skills leave unset. Version numbers are not shown because
  they do not measure review depth or maturity.}
  \label{fig:provenance}
\end{figure}

Skills are versioned individually in \texttt{metadata.version}, independently
of the repository release, and all \NumSkills{} carry a version string at this
tag.
We do not interpret the major-version distribution as a maturity measure:
non-breaking revisions can be extensive, and the field records declared
compatibility rather than review depth.
The \texttt{license} field is optional and \NumWithoutLicense{} skills leave it
unset, so for those the repository's own MIT licence is the only statement
covering them.
Authorship is recorded for \NumDeclaringAuthor{} of the \NumSkills{} skills.
We divide all skills into four groups (Figure~\ref{fig:provenance}).
Of these, \NumKDenseAuthored{} skills are attributed to us and
\NumOutsideAuthored{} to outside contributors, across
\NumDistinctOutsideAuthors{} distinct names; \NumVendoredSkills{} are
document-format skills vendored from another public skills repository, and
\NumUnattributed{} name no author.
For the last group, readers lack the per-skill attribution needed to judge
provenance.
This split means the library is not a uniformly reviewed corpus, and our own
security guidance says as much
(Appendix~\ref{app:validation}).

\subsection{Portability}

The format uses a file layout rather than a runtime interface so that different
agent hosts can discover the same skill directory.
This paper does not test a host-compatibility matrix, so portability here means
conformance to that shared layout rather than identical behaviour across
clients.
We also package the library as an Agent Plugins
1.0.0~\citep{agentplugins2026spec} bundle, with a root manifest and the
\texttt{skills/} tree.
Plugin-capable clients can discover that package, but distribution and
installation remain client-specific rather than part of either standard.

Portability still depends on the host.
Hosts differ in install paths, in discovery settings, and in which optional
header fields they support, so a skill's behaviour is not fully determined by
its own files.
The clearest case is a host-specific mapping nested under \texttt{metadata}:
at this tag \NumHostManifestBlocks{} of \NumSkills{} skills carry one.
The reference validator accepts these mappings, but the specification text
defines \texttt{metadata} as a mapping from string keys to string values.
Because the validator and specification disagree, a host may consume the
extension, ignore it or reject it.
Our contributor guide further warns that a failed eligibility check in that
mapping can hide a skill from the agent, and that writing the mapping as a
JSON string silently disables the gating and credential injection it was
meant to provide.
At this tag \NumWithCompatibility{} of \NumSkills{} skills declare a
\texttt{compatibility} string naming their environment requirements.
What they declare they may do is the \texttt{allowed-tools} field discussed
above, not this string.

\subsection{The skill directory, and a skill as it ships}

\begin{figure}[htbp]
  \centering
  \includegraphics[width=\textwidth]{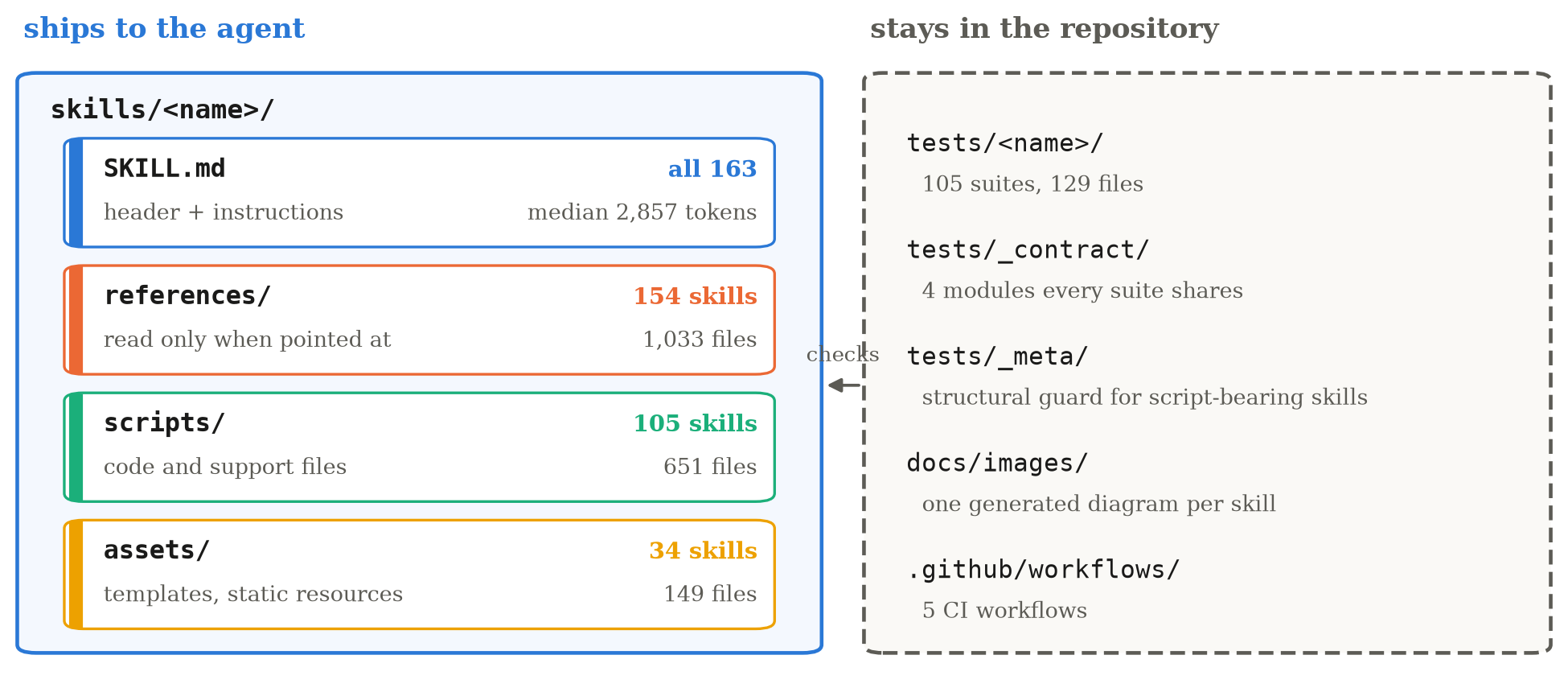}
  \caption{The skill payload and repository apparatus at tag \PinTag{}.
  Agent-loadable material lives under \texttt{skills/<name>/}; test suites,
  shared checks, generated diagrams and CI workflows live elsewhere in the
  repository. The tree follows that separation at this pin, although
  Appendix~\ref{app:validation} shows that the rule excluding in-skill tests is
  enforced only for script-bearing skills.}
  \label{fig:anatomy}
\end{figure}

Figure~\ref{fig:anatomy} separates the agent payload from the repository
infrastructure.
Everything under \texttt{skills/<name>/} ships to the agent, while test suites,
shared checks, generated diagrams and CI workflows do not.
Listing~\ref{lst:skill} shows the verbatim header and opening of one
\texttt{SKILL.md} as it ships at this tag.
Of everything in the listing, only \texttt{name} and \texttt{description} are
resident before selection.

\lstinputlisting[style=skillmd,float=htbp,caption={A skill in its shipped form:
the verbatim header and opening sections of
\texttt{skills/analytical-method-validation/SKILL.md} at tag \PinTag{}, with
the rest of the instruction body truncated. The header carries all six
permitted fields; its \texttt{name} and \texttt{description} are the only
parts counted in resident context before selection.},captionpos=b,label=lst:skill]%
{tables/skill_example.txt}

The specification and repository guidance recommend keeping instruction files
within \SkillMdLineCap{} lines and moving longer material into
\texttt{references/}.
At this tag the longest complete file, including its YAML header, is
\SkillMdMaxLines{} lines, and \SkillMdOverCap{} files exceed the limit.
The specification workflow checks all skills but would report an over-limit
file only as a warning.
The blocking structural check covers only script-bearing skills, an enforcement
gap detailed in Appendix~\ref{app:validation}.
Of the \NumWithReferences{} skills that ship reference documents, the median
ships \MedianReferencesPerSkill{} and the largest ships
\MaxReferencesPerSkill{}.

\FloatBarrier

\section{What the library covers, in detail}
\label{app:coverage}

The library's documentation groups skills under partly overlapping headings
rather than defining a single partition.
For reporting, we assigned each of the \NumSkills{} skills exactly one primary
category by its dominant intent; these assignments yield the \NumCategories{}
categories in Table~\ref{tab:taxonomy} and Figure~\ref{fig:overview}.
The assignment is ours, not the library's.
We checked that every skill appears in exactly one category, so the counts in
Table~\ref{tab:taxonomy} partition the library.
A reader who disagrees with a placement can reassign it from the skill names in
the tagged tree.
Some placements are debatable: we group skills whose purpose is to query a
named database by that function rather than by the science of the data they
return, and single-purpose analysis packages by the domain they serve rather
than the numerical method they use.

\begin{table}[htbp]
  \centering
  \small
  \caption{Primary category assignment of all \NumSkills{} skills at tag
  \PinTag{}, with how many in each category ship executable scripts and how
  many ship reference documents. The assignment is ours and is not part of the
  library.}
  \label{tab:taxonomy}
  \input{tables/taxonomy}
\end{table}

For this discussion, we group skills into four cross-cutting kinds.
They do not map directly to Table~\ref{tab:taxonomy}: each can draw on several
categories.
We give no count for them because their boundaries are ours and differ from the
partition reported in the table.

\paragraph{Package workflows.}
Many skills document how to use a specific scientific Python package:
Scanpy~\citep{wolf2018scanpy} for single-cell analysis,
Biopython~\citep{cock2009biopython} for sequence work, pysam over the
SAM/BAM formats~\citep{li2009sequence}, RDKit for cheminformatics~\citep{rdkit},
PyDESeq2~\citep{muzellec2023pydeseq2} for differential expression
following DESeq2~\citep{love2014moderated}, and
comparable coverage in materials science, quantum computing, geospatial
analysis and mass spectrometry.
These skills are not substitutes for the packages' documentation.
They focus on procedural choices that package documentation usually omits: which
defaults are wrong for particular data, the order in which operations must
occur, and which validity checks precede interpretation.

\paragraph{Data retrieval.}
A single \texttt{database-lookup} skill documents API access to
\NumDatabasesDocumented{} public databases across \NumDatabaseDomains{}
domains, with one reference file per database.
Its \NumDatabaseRefFiles{} files include \NumDatabaseGuides{} cross-cutting
guides, and its coverage is concentrated in biology and genomics
(Figure~\ref{fig:databases}).
Its instructions prioritise traceable retrieval over convenience.
The instruction body requires an explicit retrieval contract before any call,
reconciliation of retrieved counts against expected counts, and a record of
endpoints and access dates alongside every result.
It also treats API responses as untrusted input, instructing the agent not to
follow directives embedded in returned text.
Directives embedded in retrieved text form an indirect prompt-injection channel
for any retrieval-integrated agent~\citep{greshake2023injection}.
Databases whose records include user-contributed fields can expose agents to
this channel.
Dedicated skills cover resources that do not fit a generic pattern, while
multi-database wrappers extend the coverage.
Our README's headline figure of more than one hundred databases counts those
wrappers, while \NumDatabasesDocumented{} is the number we can verify by
counting files.

\paragraph{Research platforms and laboratory automation.}
This group's \NumLabPlatformSkills{} skills target systems that hold laboratory
data or execute protocols: electronic lab notebooks, sample registries, cloud
genomics
platforms, microscopy servers, liquid-handling robots and workflow engines.
Several require authentication, but the declarations do not support a clean
library-wide count of mandatory credentials.
Across the library, \NumCredentialDeclaring{} skills name one of
\NumCredentialVars{} credential-like environment variables ending in
\texttt{KEY}, \texttt{TOKEN}, \texttt{SECRET} or \texttt{PASSWORD} in either
the \texttt{compatibility} string or host-specific metadata.
This is a lexical count of declarations, not of requirements: some variables
are optional, permit higher rate limits, or provide alternatives to interactive
login.
Configuration variables such as URLs, ports and contact email addresses are
excluded.

\paragraph{Method and judgment.}
A substantial group of skills encodes procedure rather than tooling.
They cover study design and randomisation before data exist, sample-size
justification~\citep{button2013power}, test selection, assumption checking and
the interpretation of significance thresholds~\citep{wasserstein2016asa},
uncertainty and unit propagation,
hypothesis formulation bounded by available evidence, critical appraisal of
claims, and manuscript and review preparation with explicit evidence
provenance.
Compared with documentation lookup, these skills offer the kind of checklist a
careful collaborator would apply; that reliance on procedure and judgment also
makes them the hardest to evaluate.

\FloatBarrier

\section{Design rules, and what they cost}
\label{app:design}

The following four repository constraints reflect design choices, not
housekeeping rules.

\paragraph{We enforce narrow scope by declining skills.}
Our contributor guide (\texttt{AGENTS.md}) names categories of
skill that we routinely decline: general software-engineering skills, general
infrastructure with a scientific example attached, orchestrator skills that
route to other skills, and second providers for a service an existing skill
already reaches.
Our stated reason is selection pressure.
Every installed skill competes for the agent's attention on every task, so a
broad skill imposes a cost on unrelated tasks, and an
orchestrator overlaps every specialist by construction.
The resident-tier calculation in Appendix~\ref{app:method} quantifies this
selection pressure: the tier stays small because each description is short.
Whether the descriptions are specific enough for reliable selection is not
measured here.

\paragraph{Length limits are a disclosure mechanism.}
The \SkillMdLineCap{}-line limit on instruction files is intended to push long
material into \texttt{references/}, where it adds no documentation tokens to a
session until an agent follows a pointer to it.
A skill that ignored the cap would move deferred tokens into the activation
tier for every task that selects it.

\paragraph{We keep tests out of the shipped directory.}
A skill directory contains only what an agent loads, so tests, fixtures and
generated artifacts live in a parallel tree (Figure~\ref{fig:anatomy}).
This separation requires each suite to locate its skill through an explicit
anchor path rather than by traversing a relative path.
It also lets test coverage grow without enlarging what every host must scan.

\paragraph{One environment per skill, because the pins conflict.}
We do not install the skills' scientific packages into a shared
environment because some of their requirements conflict.
At this tag \NumPinnedInterpreters{} skills need an interpreter older than the
project's own, and our guidance identifies NumPy, Zarr, lxml and Transformers
conflicts between specific pairs of skills.
Test runs therefore build a throwaway environment per skill from a committed
requirements file.
The runner records \NumUnavailablePackages{} dependencies that either fall
outside its \texttt{uv}/PyPI installation path or cannot be installed in the
hosted environment.
These include GitHub-only SDKs, conda-forge builds, CUDA packages and a package
that needs a local MATLAB installation.
Those dependencies can be installed in other environments; the ledger records
a runner limitation rather than general unavailability.
Another conflict involves module names rather than versions:
\NumSharedCommonModule{} skills ship a \texttt{scripts/\_common.py}, so
collecting two skills into one interpreter would resolve \texttt{\_common} to
whichever module was imported first and silently test the wrong
file.\footnote{Our contributing guide states a smaller
figure for this; \NumSharedCommonModule{} is what the tree holds at this tag,
and is an example of the drift Section~\ref{sec:limitations} describes.}
The test runner starts a separate process for each skill to prevent this.

The library inherits the dependency conflicts of the scientific software it
documents.
Per-skill environments keep those conflicts out of the project environment.

\FloatBarrier

\section{How the library is validated}
\label{app:validation}

\begin{figure}[htbp]
  \centering
  \includegraphics[width=\textwidth]{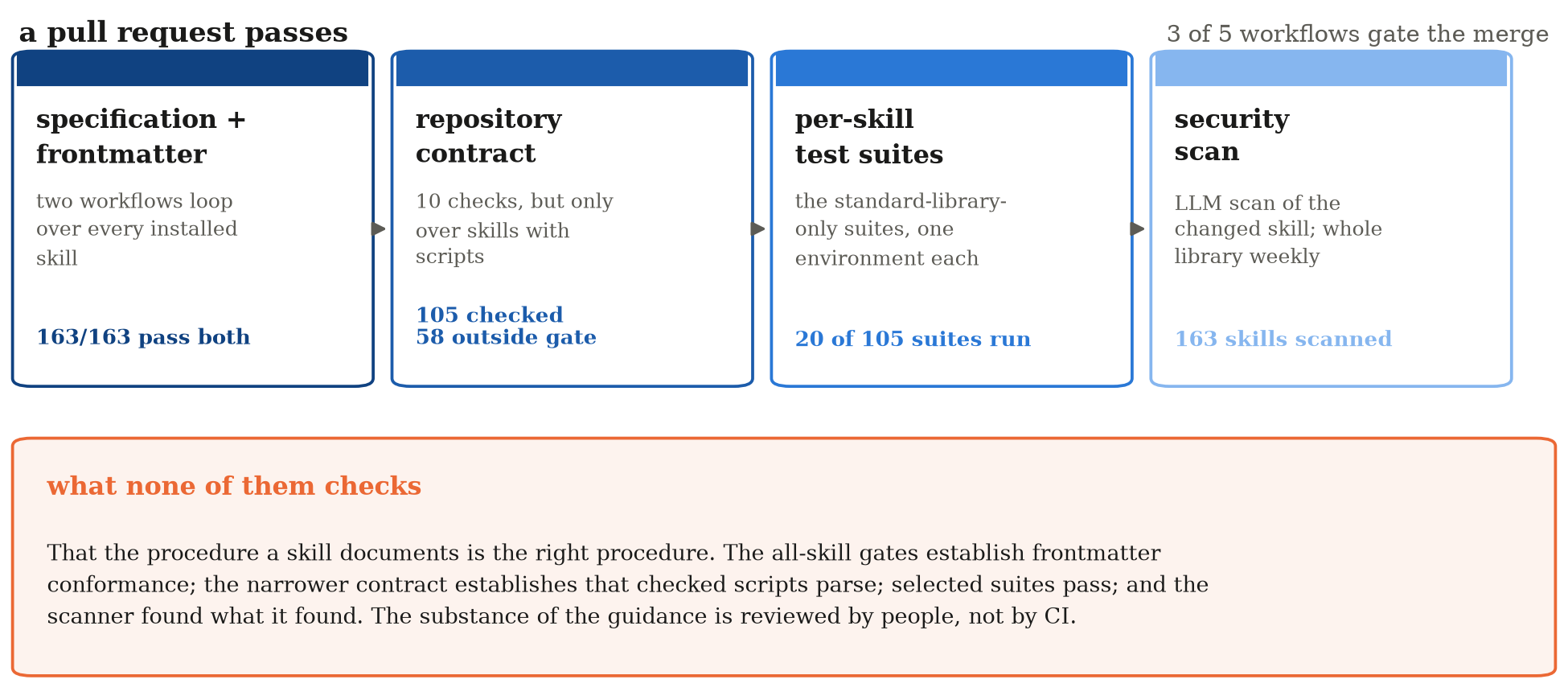}
  \caption{The gates on a relevant pull request at tag \PinTag{}.
  Frontmatter workflows loop over all \NumSkills{} skills, whereas the
  \NumStructuralChecks{}-check structural contract loops over only the
  \NumStructurallyChecked{} skills with files under \texttt{scripts/}.
  Per-skill suites exist for all \NumScriptBearing{} script-bearing skills, but
  CI runs only the \NumSuitesRunInCI{} suites whose tooling needs no scientific
  packages.
  The pull-request security scan covers changed skills; the library-wide report
  quoted in the text is the separate weekly run.
  None of these gates determines whether a documented scientific procedure is
  substantively correct.}
  \label{fig:pipeline}
\end{figure}

The term ``validated'' can refer to anything from ``someone read it'' to ``its
behaviour is tested''.
Figure~\ref{fig:pipeline} shows which checks exist and which skills they cover.
These scopes define what ``validated'' means here.

\paragraph{Structural conformance.}
The reference specification validator and a repository-specific frontmatter
check both loop over all \NumSkills{} skills.
Together they check names and descriptions, the six-field top-level set,
\texttt{metadata.version}, \texttt{allowed-tools} string syntax and the
repository's permitted host-block shape.
That host-block shape still passes the reference validator, despite the
mismatch with the specification text noted in Appendix~\ref{app:anatomy}.

The separate \texttt{tests/\_meta} suite imports a structural contract with
\NumStructuralChecks{} checks.
They cover file length, tests and bytecode outside the skill payload,
local-path resolution, Python parsing, banned dynamic execution,
standard-library name collisions, contributor-local paths and shell-script
validity, plus overlapping frontmatter checks.
Its documentation says that it spans every skill, but at this tag its loop is
over the script-bearing set.
It therefore gates \NumStructurallyChecked{} skills and leaves the other
\NumStructurallyUnchecked{} outside those checks.
An over-limit file would still reach the all-skill specification workflow, but
only as a warning.

We ran the same structural functions over the full pinned tree for this
description.
They produce \NumStructuralFindings{} findings: broken local paths in
\NumSkillsWithBrokenLocalLinks{} documentation-only skills, all outside the CI loop.
This shows that the narrower scope changes what CI catches.
Table~\ref{tab:conformance} reports each check's applicable population, actual
pull-request scope and state at the pin; it does not infer enforcement from a
passing current tree.

\begin{table}[htbp]
  \centering
  \footnotesize
  \caption{Repository checks at tag \PinTag{}. ``Checked in PR'' is the number
  of skills for which a failing result can block a relevant pull request;
  ``Holds at pin'' is our evaluation over the full population to which the rule
  applies.
  The broken paths occur in \NumSkillsWithBrokenLocalLinks{}
  documentation-only skills outside the structural contract's pull-request
  scope.}
  \label{tab:conformance}
  \input{tables/conformance}
\end{table}

\paragraph{Test coverage.}
The coverage guard asserts that every skill containing files under
\texttt{scripts/} has a corresponding test directory and an entry in the test
runner's requirements file.
At this tag that correspondence holds exactly: \NumScriptBearing{}
script-bearing skills and \NumTestSuites{} suites across \NumTestFiles{} test
files, with \NumUntestedScriptSkills{} missing suites and no orphaned suites.
We measure whether suites exist; their existence does not establish that every
helper or code path has a behavioural assertion.
The shared contract of \NumContractModules{} modules supplies reusable checks to
per-skill suites.
Its structural component parses all \NumPythonFilesInScripts{} bundled Python
files with \texttt{ast}, checks the \NumShellFilesInScripts{} shell scripts
separately, and never imports skill code.

Suite existence, execution and coverage of a particular behaviour are
different claims.
Of the \NumTestSuites{} suites, \NumSuitesRunInCI{} run on pull requests that
touch the workflow's relevant paths.
The workflow selects suites whose bundled tooling needs no scientific packages,
because the full sweep builds an environment per skill and several need CUDA, a
JDK or a local MATLAB installation unavailable on the hosted runner.
The remaining \NumSuitesLocalOnly{} are configured for the local isolated
runner, and our contributor guide instructs maintainers to run them before a
release; the pinned tree does not record whether that manual run occurred.
\NumPRWorkflows{} of the repository's \NumCIWorkflows{} workflows have
pull-request triggers: specification validation, the structural and selected
per-skill tests, and a security scan of changed skills.
The other \NumNonPRWorkflows{} run on a merge to the default branch and on a
weekly schedule.

The all-skill gates establish frontmatter conformance.
The narrower structural gate establishes parsing and path properties for
script-bearing skills.
The selected suites establish only the behaviours they assert.
None of these gates shows that following a skill's instructions produces
correct science or evaluates the substance of the guidance.

\begin{figure}[htbp]
  \centering
  \includegraphics[width=\textwidth]{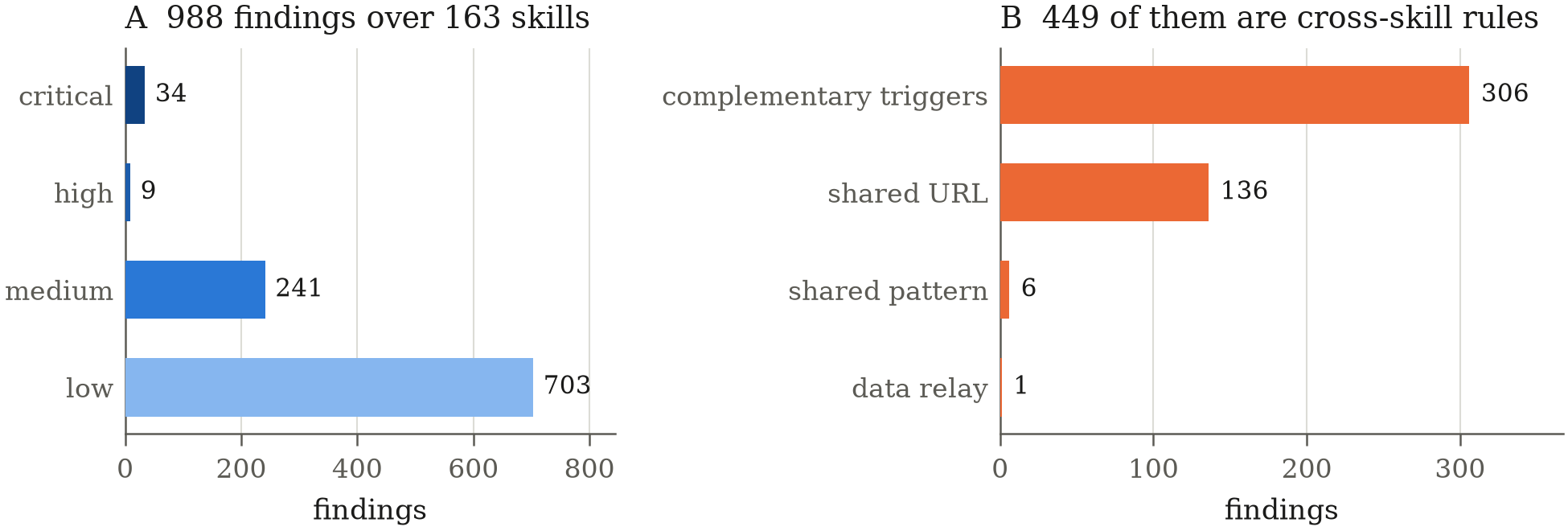}
  \caption{Published scan output at tag \PinTag{}, reported as scanner output
  rather than as a vulnerability census. \textbf{(A)} All \ScanFindings{}
  findings by severity. \textbf{(B)} The \ScanCrossSkill{} cross-skill findings
  raised against groups of skills rather than individual skills, by rule. Panel
  B isolates the share of the report produced by library-level heuristics
  rather than findings attached to individual packages.}
  \label{fig:security}
\end{figure}

\paragraph{Security scanning.}
Skills can direct consequential actions: a skill can instruct an agent to run
arbitrary code, install packages, make network requests and modify files.
We treat those capabilities as a direct risk.
The instruction file is itself a prompt-injection vector, because it is text an
agent is asked to follow~\citep{schmotz2025injection}, and susceptibility to
injected instructions can now be measured directly for tool-using
agents~\citep{debenedetti2024agentdojo}.
Surveys of public skill ecosystems report vulnerable skills at non-trivial
rates, though the reported prevalence depends heavily on the detection method
and on which repositories are
swept~\citep{liu2026wild,holzbauer2026context}.
A lifecycle analysis of the standard argues that its most serious exposures
are structural: there is no boundary between data and instructions, and the
trust decision occurs once at install time~\citep{li2026secure}.
Repository scanning cannot eliminate either exposure, and both limit what the
checks in this section can establish.
We scan every skill with a third-party scanner combining behavioural, trigger
and LLM analyzers~\citep{cisco2026skillscanner}.
Pull requests that touch relevant paths scan changed skills.
A weekly incremental scan covers the whole library, carrying previous findings
forward for unchanged skills.
We force a full rescan whenever the scanner version or the model changes, when
a maintainer triggers one, and in any case at least every thirty days.
We publish the results.
The report committed at this tag (scanner \ScannerVersion{}, model
\ScannerModel{}, generated \ScanDate{}) covers \ScanSkills{} of the
\NumSkills{} skills;
it flags \ScanSafeSkills{} as safe and records \ScanFindings{} findings:
\ScanCritical{} critical, \ScanHigh{} high, \ScanMedium{} medium,
\ScanLow{} low and \ScanInfo{} informational.
\ScanCrossSkill{} of those \ScanFindings{} are cross-skill findings, raised
against a pair or a group of skills rather than against one
(Figure~\ref{fig:security}).

The scan counts are not a vulnerability census.
\ScanNotSafe{} skills are not flagged safe, and \ScanCriticalSkills{} carry at
least one critical finding.
Of the \ScanCritical{} critical findings, \ScanCriticalExfilRules{} come from
the \ScanExfilRuleCount{} rules that fire when one package both reads an
environment variable and makes a network call.
Of the \ScanCriticalSkills{} skills with a critical finding,
\ScanCriticalCredentialed{} name a credential-like environment variable in
their own header.
Our published triage document (\texttt{docs/security-triage.md})
identifies exactly this pattern, in which a skill reads its own API key and
calls its own service, as a systematic false positive for those rules.
Our \texttt{SECURITY.md} accordingly asks readers to verify a finding against
the skill before acting on it.

The triage predates the report committed at this tag.
It covers the scan of \TriageScanDate{}, with \TriageScanSkills{} skills and
\TriageScanFindings{} findings, rather than the current report's
\ScanSkills{} skills and \ScanFindings{} findings.
Its verdict that no critical finding survived verification therefore applies to
the \TriageScanCritical{} critical findings, not to the \ScanCritical{} reported
here; the remaining findings are untriaged at this pin.
We count only those \ScanExfilRuleCount{} rules as sources of this systematic
false positive.
A fourth rule also has ``exfiltration'' in its name, but the same triage
document records one of its findings as genuine and already fixed, so we do
not fold that rule into the false-positive count.

The cross-skill rules also fire on structural coincidence.
The dominant rule flags skills with complementary trigger conditions, and one
finding treats nearly every data-reading skill at once as a potential relay
chain.
We report the published, checkable counts but do not convert them into a
vulnerability count, which the report does not support.
The largest ecosystem survey finds that scanner verdicts used alone
substantially overestimate maliciousness, while checking a flagged skill
against its repository removes the great majority of
suspicions~\citep{holzbauer2026context}.
Instead, the report documents our security practices.
We publish scan output, describe a triage process and a route for contesting
findings, advise users not to install every skill, and tell them to read a
skill before trusting it.

\FloatBarrier

\section{What the skills claim, and what they decline}
\label{app:scope}

Wrong answers delivered with confidence can cause harm in several domains
covered here: clinical decision support, treatment documentation, diagnostic
imaging, regulatory evidence and laboratory animal welfare.
Our category descriptions in \texttt{README.md} place whole areas outside the
scope of decision-making.
The regulatory and standards category describes its output as
``prepared for qualified review, never a certification, accreditation, or
method-release decision''.
Imaging and pathology work is described as research-only.
The healthcare-AI category covers retrospective validation and is ``not
patient-specific diagnosis, treatment, alarms, or deployment decisions''.
The clinical category carries no comparable disclaimer of its own.
Instead, it names narrow deliverables: aggregate decision-support evaluation,
draft report structures that must stay bound to their source facts, and
formatting of treatment decisions a clinician has already written.
Individual skills also state limits in their instructions.
The clinical reporting skill, for instance, marks its output as a draft not for
clinical use, requires a verified source-fact manifest before it produces any
output, and instructs the agent to stop when source support or qualified review
is missing.
The standards-readiness skill states that it is not for compliance,
certification or accreditation decisions and declines to reproduce clause text
from the standards it prepares evidence against.

\begin{figure}[htbp]
  \centering
  \includegraphics[width=\textwidth]{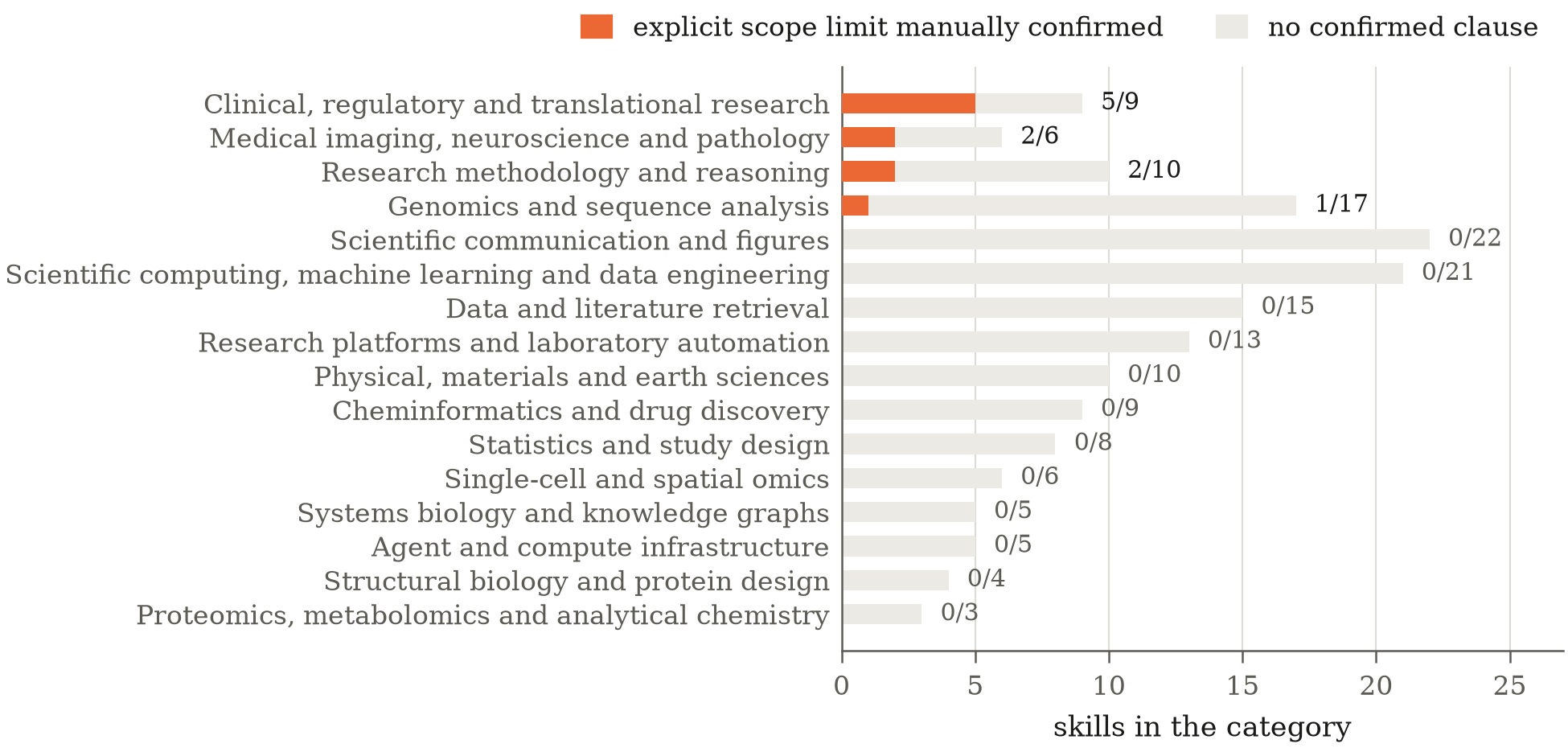}
  \caption{Where a manually confirmed scope-limiting clause appears in the file
  an agent reads, by primary category at tag \PinTag{}. Candidate files come
  from a lexical search, but each filled segment corresponds to an exact clause
  in that file. The confirmed set may miss different wording,
  and the figure shows concentration rather than a pass mark.}
  \label{fig:scope}
\end{figure}

These limits are appropriate to the stated uses and should be reported without
widening or understating them.
In the instruction files loaded on activation, however, scope clauses are
concentrated rather than uniform (Figure~\ref{fig:scope}).
We first searched every instruction file with a broad lexical rule.
We then manually adjudicated every reported positive, counting a skill only
when we could point to an exact supporting clause in its instruction file.
This procedure confirms an explicit scope limit in \NumScopeLimited{} of
\NumSkills{} instruction files.
The limits are concentrated where the stakes are highest:
\ScopeClinicalLimited{} of the \ScopeClinicalTotal{} skills in the clinical,
regulatory and translational category, and \ScopeImagingLimited{} of the
\ScopeImagingTotal{} in medical imaging, neuroscience and pathology.
The confirmed set is a lower bound because the lexical rule may miss different
wording.
Manual adjudication prevents a keyword mentioned only in passing from counting
as a positive.
Our README frames categories as research-only, but only some skills in those
categories restate the caveat.
The agent loads the instruction file, not the README.
A user or host installing a single skill by name need not encounter the
category-level framing at all.

A paper that reports using a clinical or regulatory skill should state the
limits in that specific skill because skills do not uniformly inherit the
library-level limits.

\FloatBarrier

\section{Routing sensitivity}
\label{app:sensitivity}

The routing measurement in Section~\ref{sec:routing} depends on how a
description is turned into a bag of terms, and that choice is ours.
The reported preprocessing lower-cases, keeps alphanumeric tokens of at least
\RouteMinTokenLen{} characters, and removes a \RouteStopWords{}-word
function-word list.
We tested whether the finding depends on that list by recomputing the whole
pairwise ranking under three alternatives: no stop list at all, a
minimum token length of four, and a stop list extended with domain filler
(\texttt{skill}, \texttt{data}, \texttt{analysis}, \texttt{workflow} and
similar).
Under each alternative, at least \RouteSensitivityWorst{} of the original
\RouteSensitivityN{} closest pairs remain in that alternative's
top-\RouteSensitivityN{} set.
The named collisions in Figure~\ref{fig:routing}B are therefore a property of
the descriptions rather than of the filter applied to them.

We also compare the result with the security scanner's
complementary-description rule.
The rule identifies a largely different set of pairs.
It flags \RouteScannerPairs{} pairs with a mean cosine of
\RouteScannerMeanCos{}, compared with \RouteScannerUnflaggedCos{} for unflagged
pairs.
Only \RouteScannerTopTwenty{} of the twenty closest pairs are among those it
flags.
Every input to this measurement is a \texttt{description} field in the tagged
tree, and the preprocessing is fully specified above, so a reader who prefers a
different scheme can substitute it and recompute the ranking, the
\RoutePointerEdges{} authored pointers and these comparisons without our
code.

\section{Estimating the installed base}
\label{app:base}

Every other number in this paper is computed from the pinned tree and can be
recomputed by anyone who clones it.
We keep the numbers in this appendix separate because they are not computed
from the pinned tree.
GitHub's repository traffic endpoint is readable only by accounts with push
access.
It returns a rolling \TrafficWindowDays{}-day window that GitHub does not
archive.
The result is therefore a dated snapshot that a reviewer cannot reproduce and
that we cannot reconstruct after the window passes.
We report it because the alternative is a company-reported user count with no
stated method.
We state the estimate's assumptions so that readers can scrutinise them.

From \TrafficWindowStart{} to \TrafficWindowEnd{}, the repository recorded
\TrafficClones{} clones from \TrafficCloners{} distinct cloners and
\TrafficViews{} page views from \TrafficVisitors{} distinct visitors.
Clone volume is not a count of people.
Active cloners generated an average of roughly \BaseClonesPerActiveDay{} clones
per active day.
That repetition indicates continuous integration and mirroring rather than
one-time installation.
We therefore use the distinct-actor count to estimate the population from which
the observed actors were sampled.

A single window observes only actors that cloned during that period.
Summing the per-day unique counts gives $D = \BaseDailyUniqueSum{}$.
This exceeds the window-level unique count $P = \BaseFloor{}$ because an actor
active on several days is counted once in $P$ but repeatedly in $D$.
Their ratio, $k = D/P = \BaseActiveDays{}$, is the mean number of active days
among observed cloners.
We use that conditional mean to estimate the sampling rate.
We model each actor in a pool of size $N$ as cloning independently on each day
with probability $p$.
An actor is then observed at least once with probability
$q = 1-(1-p)^{\TrafficWindowDays{}}$.
Among observed actors, the zero-truncated mean number of active days is
$\TrafficWindowDays{}\,p/q$.
Equating this expression with the measured $k$ gives
$p = \BaseDailyProb{}$ and $q = \BaseObservedProb{}$.
Under the model, a fortnight observes fewer than half of the pool.
The estimate is therefore $N = P/q \approx \BaseEstimate{}$.

Three limitations constrain what this estimate supports.
First, the model assumes a homogeneous pool, whereas a build server may clone
daily and a person installing the library may clone once.
This heterogeneity inflates $k$, the mean active days among observed cloners,
which inflates the implied observation rate $q$ and so reduces the estimated
pool $N$.
The resulting bias lowers the estimate and is therefore conservative given the
authors' interest in a larger value.
Second, GitHub's unit is a distinct cloning actor, not a person.
Shared egress can collapse a department into one actor, while one person's
laptop and CI can count as two, and we cannot determine the net direction of
those effects.
Third, the estimate describes clients that fetch the repository, not agent
sessions in which a skill was selected and used.
No public signal supplies the latter quantity.

Two independent quantities bracket the result.
The \BaseFloor{} distinct cloners observed directly provide a floor that
requires no model.
Applying the same estimator to page visitors gives a higher estimate of
\BaseViewEstimate{} because visiting is a weaker commitment than cloning.
The repository's \Stars{} stars are independent of both estimates and
accumulate over its lifetime.
The star count is within \BaseStarsDiff{}\% of the clone-pool estimate, so it is
a check rather than an input.
This agreement does not confirm scientific use because both signals measure
interest in a repository, not use of a skill.

\FloatBarrier

\end{document}

%% file: numbers.tex
\newcommand{\AllSkillMdResident}{482{,}506}
\newcommand{\BaseActiveDays}{1.29}
\newcommand{\BaseClonesPerActiveDay}{8}
\newcommand{\BaseDailyProb}{0.041}
\newcommand{\BaseDailyUniqueSum}{20{,}234}
\newcommand{\BaseEstimate}{35{,}690}
\newcommand{\BaseEstimateRounded}{36{,}000}
\newcommand{\BaseFloor}{15{,}682}
\newcommand{\BaseObservedProb}{0.44}
\newcommand{\BaseStarsDiff}{6}
\newcommand{\BaseViewEstimate}{51{,}089}
\newcommand{\BashBlocks}{625}
\newcommand{\ComboAllThree}{31}
\newcommand{\ComboReferencesOnly}{52}
\newcommand{\ComboReferencesScripts}{68}
\newcommand{\ComboSkillMdOnly}{3}
\newcommand{\CompositionChords}{147}
\newcommand{\CompositionCoreSkills}{44}
\newcommand{\CompositionMinShared}{3}
\newcommand{\CompositionMinWorkflows}{3}
\newcommand{\CorpusWindows}{14.8}
\newcommand{\DatabasesKeyed}{18}
\newcommand{\DatabasesOpen}{57}
\newcommand{\DatabasesRestricted}{3}
\newcommand{\DistinctHeadings}{844}
\newcommand{\DocTokensTotal}{2{,}963{,}180}
\newcommand{\FitAllSkillMd}{89}
\newcommand{\FitWholeCorpus}{38}
\newcommand{\FracDeferred}{83.7}
\newcommand{\HeaviestTaskTokens}{231{,}263}
\newcommand{\MaxReferenceFileTokens}{189{,}746}
\newcommand{\MaxReferencesPerSkill}{80}
\newcommand{\MedianCodeBlocks}{8}
\newcommand{\MedianHeadingsPerSkill}{10}
\newcommand{\MedianReferenceFileTokens}{2{,}000}
\newcommand{\MedianReferencesPerSkill}{6}
\newcommand{\MedianScriptsPerSkill}{4}
\newcommand{\MedianTaskFracWindow}{8.6}
\newcommand{\MedianTaskTokens}{17{,}103}
\newcommand{\MedianTaskWithReferences}{29{,}440}
\newcommand{\MostComposedSkill}{scientific-visualization}
\newcommand{\MostComposedSkillWorkflows}{27}

\newcommand{\NumCIWorkflows}{5}
\newcommand{\NumCategories}{16}
\newcommand{\NumContractModules}{4}
\newcommand{\NumCredentialDeclaring}{29}
\newcommand{\NumCredentialVars}{27}
\newcommand{\NumDatabaseDomains}{9}
\newcommand{\NumDatabaseGuides}{2}
\newcommand{\NumDatabaseRefFiles}{80}
\newcommand{\NumDatabasesDocumented}{78}
\newcommand{\NumDeclaringAuthor}{159}
\newcommand{\NumDeclaringTools}{100}
\newcommand{\NumDistinctOutsideAuthors}{17}
\newcommand{\NumDocWorkflows}{46}
\newcommand{\NumExternallyAuthored}{32}
\newcommand{\NumHostManifestBlocks}{25}
\newcommand{\NumKDenseAuthored}{131}
\newcommand{\NumLabPlatformSkills}{13}

\newcommand{\NumNonPRWorkflows}{2}
\newcommand{\NumNotDeclaringTools}{63}
\newcommand{\NumOutsideAuthored}{24}
\newcommand{\NumPRWorkflows}{3}
\newcommand{\NumPinnedInterpreters}{10}
\newcommand{\NumPythonFilesInScripts}{527}
\newcommand{\NumReferenceFiles}{1{,}033}
\newcommand{\NumScopeLimited}{10}
\newcommand{\NumScriptBearing}{105}
\newcommand{\NumScriptFiles}{651}
\newcommand{\NumSharedCommonModule}{40}
\newcommand{\NumShellFilesInScripts}{2}
\newcommand{\NumSkills}{163}
\newcommand{\NumSkillsWithBrokenLocalLinks}{2}
\newcommand{\NumStructuralChecks}{10}
\newcommand{\NumStructuralFindings}{2}
\newcommand{\NumStructurallyChecked}{105}
\newcommand{\NumStructurallyUnchecked}{58}
\newcommand{\NumSuitesLocalOnly}{85}
\newcommand{\NumSuitesRunInCI}{20}
\newcommand{\NumSupportFilesInScripts}{122}
\newcommand{\NumTestFiles}{129}
\newcommand{\NumTestSuites}{105}
\newcommand{\NumUnattributed}{4}
\newcommand{\NumUnavailablePackages}{18}
\newcommand{\NumUntestedScriptSkills}{0}
\newcommand{\NumVendoredSkills}{4}

\newcommand{\NumWithAssets}{34}
\newcommand{\NumWithCompatibility}{116}
\newcommand{\NumWithReferences}{154}
\newcommand{\NumWithScripts}{105}

\newcommand{\NumWithoutLicense}{4}
\newcommand{\PinDate}{2026-08-29}
\newcommand{\PinSha}{f6fcafeb1cc8c82eca0160a18bc41c38427b8e0f}

\newcommand{\PinTag}{v2.65.0}
\newcommand{\PythonBlocks}{760}

\newcommand{\ReleaseWindowDays}{311}
\newcommand{\ReleasesInWindow}{100}
\newcommand{\ReleasesPerMonth}{9.7}
\newcommand{\ResidentFracWindow}{7.1}
\newcommand{\RouteClosestA}{datamol}
\newcommand{\RouteClosestB}{rdkit}
\newcommand{\RouteClosestCos}{0.40}
\newcommand{\RouteEnrichment}{14}
\newcommand{\RouteGuardedTopFifty}{26}
\newcommand{\RouteGuardedTopHundred}{33}
\newcommand{\RouteGuardedTopTwenty}{17}

\newcommand{\RouteMeanCosNoPointer}{0.013}
\newcommand{\RouteMeanCosPointer}{0.185}
\newcommand{\RouteMedianNN}{0.15}

\newcommand{\RouteMinTokenLen}{3}
\newcommand{\RoutePNinetyNN}{0.26}
\newcommand{\RoutePairsTotal}{13{,}203}
\newcommand{\RoutePointerEdges}{56}
\newcommand{\RoutePointerPairs}{44}
\newcommand{\RoutePointerReciprocal}{12}
\newcommand{\RoutePointerSources}{33}

\newcommand{\RouteScannerMeanCos}{0.020}
\newcommand{\RouteScannerPairs}{301}
\newcommand{\RouteScannerTopTwenty}{0}
\newcommand{\RouteScannerUnflaggedCos}{0.013}
\newcommand{\RouteSensitivityN}{10}
\newcommand{\RouteSensitivityWorst}{9}
\newcommand{\RouteStopWords}{27}
\newcommand{\RouteUnguardedA}{infographics}
\newcommand{\RouteUnguardedB}{scientific-schematics}
\newcommand{\RouteUnguardedCos}{0.36}
\newcommand{\ScanCritical}{34}
\newcommand{\ScanCriticalCredentialed}{8}
\newcommand{\ScanCriticalExfilRules}{28}
\newcommand{\ScanCriticalSkills}{11}
\newcommand{\ScanCrossSkill}{449}

\newcommand{\ScanDate}{2026-08-24}
\newcommand{\ScanExfilRuleCount}{3}
\newcommand{\ScanFindings}{988}
\newcommand{\ScanHigh}{9}
\newcommand{\ScanInfo}{1}
\newcommand{\ScanLow}{703}
\newcommand{\ScanMedium}{241}
\newcommand{\ScanNotSafe}{16}
\newcommand{\ScanSafeSkills}{147}
\newcommand{\ScanSkills}{163}

\newcommand{\ScannerModel}{claude-opus-5}
\newcommand{\ScannerVersion}{2.0.13}
\newcommand{\ScopeClinicalLimited}{5}
\newcommand{\ScopeClinicalTotal}{9}
\newcommand{\ScopeImagingLimited}{2}
\newcommand{\ScopeImagingTotal}{6}
\newcommand{\SecondComposedSkill}{scientific-writing}
\newcommand{\SecondComposedSkillWorkflows}{26}
\newcommand{\SkillMdLineCap}{500}
\newcommand{\SkillMdMaxLines}{496}
\newcommand{\SkillMdMedianLines}{285}
\newcommand{\SkillMdOverCap}{0}
\newcommand{\SkillsWithCaveats}{84}
\newcommand{\SkillsWithDeferredSection}{143}
\newcommand{\SkillsWithProcedure}{126}
\newcommand{\SkillsWithSelectionCue}{104}
\newcommand{\Stars}{37{,}905}
\newcommand{\StarsDate}{2026-08-29}
\newcommand{\TierOneAsFracCorpus}{0.48}
\newcommand{\TierOneAsFracTierTwo}{2.9}

\newcommand{\TierOneMedianTokens}{67}
\newcommand{\TierOneTokens}{14{,}246}

\newcommand{\TierThreeTokens}{2{,}480{,}674}
\newcommand{\TierTwoMedianTokens}{2{,}857}
\newcommand{\TierTwoTokens}{482{,}506}
\newcommand{\Tokenizer}{o200k\_base}
\newcommand{\ToolBash}{94}
\newcommand{\ToolEdit}{78}
\newcommand{\ToolRead}{99}
\newcommand{\ToolWrite}{92}
\newcommand{\TotalCodeBlocks}{1{,}489}
\newcommand{\TrafficCloners}{15{,}682}
\newcommand{\TrafficClones}{161{,}210}
\newcommand{\TrafficViews}{61{,}901}
\newcommand{\TrafficVisitors}{21{,}268}
\newcommand{\TrafficWindowDays}{14}
\newcommand{\TrafficWindowEnd}{2026-08-27}
\newcommand{\TrafficWindowStart}{2026-08-14}
\newcommand{\TriageScanCritical}{33}
\newcommand{\TriageScanDate}{2026-07-27}
\newcommand{\TriageScanFindings}{817}
\newcommand{\TriageScanSkills}{154}
\newcommand{\UntaggedBlocks}{67}
\newcommand{\WindowTokens}{200{,}000}
\newcommand{\WorkflowAffordAll}{17}
\newcommand{\WorkflowAffordMedian}{85}
\newcommand{\WorkflowAffordMin}{35}
\newcommand{\WorkflowAffordQOne}{61}
\newcommand{\WorkflowAffordUnderHalf}{5}

\newcommand{\WorkflowCategoriesMedian}{4}

\newcommand{\WorkflowMaxFull}{466{,}131}
\newcommand{\WorkflowMaxInstruction}{62{,}476}
\newcommand{\WorkflowMaxSkills}{16}
\newcommand{\WorkflowMedianFull}{225{,}498}
\newcommand{\WorkflowMedianFullPct}{112.7}
\newcommand{\WorkflowMedianInstruction}{47{,}706}
\newcommand{\WorkflowMedianInstructionPct}{23.9}
\newcommand{\WorkflowMedianSkills}{10}
\newcommand{\WorkflowMinSkills}{5}
\newcommand{\WorkflowOverWindow}{29}
\newcommand{\WorkflowOverWindowInstruction}{0}

\newcommand{\WorkflowSkillsUsed}{162}
\newcommand{\WorkflowSkillsUsedOnce}{81}
\newcommand{\WorkflowSlots}{472}

%% file: tables/taxonomy.tex
% Generated by `uv run python -m sas_paper.tables`. Do not edit.
\begin{tabular}{lrrr}
\toprule
Category & Skills & With scripts & With references \\
\midrule
Scientific communication and figures & 22 & 18 & 18 \\
Scientific computing, machine learning and data engineering & 21 & 10 & 21 \\
Genomics and sequence analysis & 17 & 12 & 16 \\
Data and literature retrieval & 15 & 8 & 12 \\
Research platforms and laboratory automation & 13 & 9 & 13 \\
Physical, materials and earth sciences & 10 & 6 & 10 \\
Research methodology and reasoning & 10 & 6 & 10 \\
Cheminformatics and drug discovery & 9 & 5 & 9 \\
Clinical, regulatory and translational research & 9 & 8 & 9 \\
Statistics and study design & 8 & 7 & 8 \\
Medical imaging, neuroscience and pathology & 6 & 5 & 6 \\
Single-cell and spatial omics & 6 & 2 & 6 \\
Agent and compute infrastructure & 5 & 2 & 5 \\
Systems biology and knowledge graphs & 5 & 4 & 4 \\
Structural biology and protein design & 4 & 0 & 4 \\
Proteomics, metabolomics and analytical chemistry & 3 & 3 & 3 \\
\midrule
\textbf{Total} & \textbf{163} & \textbf{105} & \textbf{154} \\
\bottomrule
\end{tabular}

%% file: tables/conformance.tex
% Generated by `uv run python -m sas_paper.tables`. Do not edit.
\begin{tabular}{p{0.55\textwidth}rrr}
\toprule
Rule & Applies & Checked in PR & Holds at pin \\
\midrule
\multicolumn{4}{l}{\emph{Fully enforced for its scope}} \\
\quad reference specification validator & 163 & 163 & 163 \\
\quad repository frontmatter additions & 163 & 163 & 163 \\
\quad bundled Python files parse & 105 & 105 & 105 \\
\quad no banned dynamic-execution calls & 105 & 105 & 105 \\
\quad no top-level standard-library shadowing & 105 & 105 & 105 \\
\quad bundled shell scripts parse and are executable & 2 & 2 & 2 \\
\quad script-bearing skill has a test suite & 105 & 105 & 105 \\
\quad script-bearing skill has a requirements entry & 105 & 105 & 105 \\
\addlinespace
\multicolumn{4}{l}{\emph{Enforced only for script-bearing skills}} \\
\quad SKILL.md within the 500-line limit & 163 & 105 & 163 \\
\quad no tests shipped inside a skill directory & 163 & 105 & 163 \\
\quad no compiled bytecode shipped & 163 & 105 & 163 \\
\quad documented local paths resolve & 163 & 105 & 161 \\
\quad no contributor-local paths & 163 & 105 & 163 \\
\addlinespace
\multicolumn{4}{l}{\emph{Convention, not a pull-request gate}} \\
\quad generated workflow diagram present & 163 & 0 & 162 \\
\quad listed in the documentation catalogue & 163 & 0 & 162 \\
\bottomrule
\end{tabular}

%% file: main.bbl
\begin{thebibliography}{57}
\providecommand{\natexlab}[1]{#1}
\providecommand{\url}[1]{\texttt{#1}}
\expandafter\ifx\csname urlstyle\endcsname\relax
  \providecommand{\doi}[1]{doi: #1}\else
  \providecommand{\doi}{doi: \begingroup \urlstyle{rm}\Url}\fi

\bibitem[Benjamini and Hochberg(1995)]{benjamini1995controlling}
Yoav Benjamini and Yosef Hochberg.
\newblock {Controlling the False Discovery Rate: A Practical and Powerful
  Approach to Multiple Testing}.
\newblock \emph{Journal of the Royal Statistical Society Series B: Statistical
  Methodology}, 57\penalty0 (1):\penalty0 289–300, January 1995.
\newblock ISSN 1467-9868.
\newblock \doi{10.1111/j.2517-6161.1995.tb02031.x}.

\bibitem[Lazic(2010)]{lazic2010pseudoreplication}
Stanley~E Lazic.
\newblock {The problem of pseudoreplication in neuroscientific studies: is it
  affecting your analysis?}
\newblock \emph{BMC Neuroscience}, 11\penalty0 (1):\penalty0 5, January 2010.
\newblock ISSN 1471-2202.
\newblock \doi{10.1186/1471-2202-11-5}.

\bibitem[Leek et~al.(2010)Leek, Scharpf, Bravo, Simcha, Langmead, Johnson,
  Geman, Baggerly, and Irizarry]{leek2010tackling}
Jeffrey~T. Leek, Robert~B. Scharpf, Héctor~Corrada Bravo, David Simcha,
  Benjamin Langmead, W.~Evan Johnson, Donald Geman, Keith Baggerly, and
  Rafael~A. Irizarry.
\newblock {Tackling the widespread and critical impact of batch effects in
  high-throughput data}.
\newblock \emph{Nature Reviews Genetics}, 11\penalty0 (10):\penalty0 733–739,
  September 2010.
\newblock ISSN 1471-0064.
\newblock \doi{10.1038/nrg2825}.

\bibitem[{UCSC Genome Browser}(2026)]{ucsc2026coordinates}
{UCSC Genome Browser}.
\newblock Format and coordinate conventions.
\newblock \url{https://genome.ucsc.edu/FAQ/FAQformat.html}, 2026.
\newblock Accessed 2026-09-02.

\bibitem[Ziemann et~al.(2016)Ziemann, Eren, and El-Osta]{ziemann2016gene}
Mark Ziemann, Yotam Eren, and Assam El-Osta.
\newblock {Gene name errors are widespread in the scientific literature}.
\newblock \emph{Genome Biology}, 17\penalty0 (1):\penalty0 177, August 2016.
\newblock ISSN 1474-760X.
\newblock \doi{10.1186/s13059-016-1044-7}.

\bibitem[Brueckner et~al.(2026)Brueckner, Patel, He, and
  Kassis]{brueckner2026kbench}
Aubrey Brueckner, Darshil Patel, Yuhuan He, and Timothy Kassis.
\newblock {K-Bench: measuring model performance on real scientific agent
  requests}, 2026.
\newblock arXiv:2608.21601 [cs.AI].

\bibitem[Percie~du Sert et~al.(2020)Percie~du Sert, Hurst, Ahluwalia, Alam,
  Avey, Baker, Browne, Clark, Cuthill, Dirnagl, Emerson, Garner, Holgate,
  Howells, Karp, Lazic, Lidster, MacCallum, Macleod, Pearl, Petersen, Rawle,
  Reynolds, Rooney, Sena, Silberberg, Steckler, and Würbel]{percie2020arrive}
Nathalie Percie~du Sert, Viki Hurst, Amrita Ahluwalia, Sabina Alam, Marc~T.
  Avey, Monya Baker, William~J. Browne, Alejandra Clark, Innes~C. Cuthill,
  Ulrich Dirnagl, Michael Emerson, Paul Garner, Stephen~T. Holgate, David~W.
  Howells, Natasha~A. Karp, Stanley~E. Lazic, Katie Lidster, Catriona~J.
  MacCallum, Malcolm Macleod, Esther~J. Pearl, Ole~H. Petersen, Frances Rawle,
  Penny Reynolds, Kieron Rooney, Emily~S. Sena, Shai~D. Silberberg, Thomas
  Steckler, and Hanno Würbel.
\newblock {The ARRIVE guidelines 2.0: Updated guidelines for reporting animal
  research}.
\newblock \emph{PLOS Biology}, 18\penalty0 (7):\penalty0 e3000410, July 2020.
\newblock ISSN 1545-7885.
\newblock \doi{10.1371/journal.pbio.3000410}.

\bibitem[Bustin et~al.(2009)Bustin, Benes, Garson, Hellemans, Huggett, Kubista,
  Mueller, Nolan, Pfaffl, Shipley, Vandesompele, and Wittwer]{bustin2009miqe}
Stephen~A Bustin, Vladimir Benes, Jeremy~A Garson, Jan Hellemans, Jim Huggett,
  Mikael Kubista, Reinhold Mueller, Tania Nolan, Michael~W Pfaffl, Gregory~L
  Shipley, Jo~Vandesompele, and Carl~T Wittwer.
\newblock {The MIQE Guidelines: Minimum Information for Publication of
  Quantitative Real-Time PCR Experiments}.
\newblock \emph{Clinical Chemistry}, 55\penalty0 (4):\penalty0 611–622, April
  2009.
\newblock ISSN 1530-8561.
\newblock \doi{10.1373/clinchem.2008.112797}.

\bibitem[Makin and Orban~de Xivry(2019)]{makin2019ten}
Tamar~R Makin and Jean-Jacques Orban~de Xivry.
\newblock {Ten common statistical mistakes to watch out for when writing or
  reviewing a manuscript}.
\newblock \emph{eLife}, 8:\penalty0 e48175, October 2019.
\newblock ISSN 2050-084X.
\newblock \doi{10.7554/elife.48175}.

\bibitem[Agent Skills()]{agentskills2026spec}
Agent Skills.
\newblock Agent skills specification.
\newblock \url{https://agentskills.io/specification}, 2026.
\newblock Accessed 2026-09-02.

\bibitem[{K-Dense Inc.}(2026)]{kdense2026skills}
{K-Dense Inc.}
\newblock Scientific agent skills.
\newblock \url{https://github.com/K-Dense-AI/scientific-agent-skills}, 2026.
\newblock Version 2.65.0, commit f6fcafe; accessed 2026-09-02.

\bibitem[Boiko et~al.(2023)Boiko, MacKnight, Kline, and
  Gomes]{boiko2023autonomous}
Daniil~A. Boiko, Robert MacKnight, Ben Kline, and Gabe Gomes.
\newblock {Autonomous chemical research with large language models}.
\newblock \emph{Nature}, 624\penalty0 (7992):\penalty0 570–578, December
  2023.
\newblock ISSN 1476-4687.
\newblock \doi{10.1038/s41586-023-06792-0}.

\bibitem[Li et~al.(2025)Li, Agarwal, Zhou, Gopinath, and
  Kassis]{li2025kdenseanalyst}
Orion Li, Vinayak Agarwal, Summer Zhou, Ashwin Gopinath, and Timothy Kassis.
\newblock {K-Dense Analyst: Towards Fully Automated Scientific Analysis}, 2025.
\newblock arXiv:2508.07043 [cs.AI].

\bibitem[Lu et~al.(2024)Lu, Lu, Lange, Foerster, Clune, and Ha]{lu2024ai}
Chris Lu, Cong Lu, Robert~Tjarko Lange, Jakob Foerster, Jeff Clune, and David
  Ha.
\newblock {The AI Scientist: Towards Fully Automated Open-Ended Scientific
  Discovery}, 2024.
\newblock arXiv:2408.06292 [cs.AI].

\bibitem[{MIMS Harvard}(2026)]{tooluniverse2026}
{MIMS Harvard}.
\newblock Tooluniverse ai agent skills.
\newblock
  \url{https://zitniklab.hms.harvard.edu/ToolUniverse/guide/skills_showcase.html},
  2026.
\newblock Accessed 2026-09-02.

\bibitem[Jimenez et~al.(2023)Jimenez, Yang, Wettig, Yao, Pei, Press, and
  Narasimhan]{jimenez2024swebench}
Carlos~E. Jimenez, John Yang, Alexander Wettig, Shunyu Yao, Kexin Pei, Ofir
  Press, and Karthik Narasimhan.
\newblock {SWE-bench: Can Language Models Resolve Real-World GitHub Issues?},
  2023.
\newblock arXiv:2310.06770 [cs.CL].

\bibitem[Tian et~al.(2024)Tian, Gao, Zhang, Chen, Fan, Guo, Haas, Ji,
  Krongchon, Li, Liu, Luo, Ma, Tong, Trinh, Tian, Wang, Wu, Xiong, Yin, Zhu,
  Lieret, Lu, Liu, Du, Tao, Press, Callan, Huerta, and Peng]{tian2024scicode}
Minyang Tian, Luyu Gao, Shizhuo~Dylan Zhang, Xinan Chen, Cunwei Fan, Xuefei
  Guo, Roland Haas, Pan Ji, Kittithat Krongchon, Yao Li, Shengyan Liu, Di~Luo,
  Yutao Ma, Hao Tong, Kha Trinh, Chenyu Tian, Zihan Wang, Bohao Wu, Yanyu
  Xiong, Shengzhu Yin, Minhui Zhu, Kilian Lieret, Yanxin Lu, Genglin Liu,
  Yufeng Du, Tianhua Tao, Ofir Press, Jamie Callan, Eliu Huerta, and Hao Peng.
\newblock {SciCode: A Research Coding Benchmark Curated by Scientists}, 2024.
\newblock arXiv:2407.13168 [cs.AI].

\bibitem[Chen et~al.(2024)Chen, Chen, Ning, Zhang, Wang, Yu, Li, Liao, Wei, Lu,
  Dey, Xue, Baker, Burns, Adu-Ampratwum, Huang, Ning, Gao, Su, and
  Sun]{chen2024scienceagentbench}
Ziru Chen, Shijie Chen, Yuting Ning, Qianheng Zhang, Boshi Wang, Botao Yu,
  Yifei Li, Zeyi Liao, Chen Wei, Zitong Lu, Vishal Dey, Mingyi Xue, Frazier~N.
  Baker, Benjamin Burns, Daniel Adu-Ampratwum, Xuhui Huang, Xia Ning, Song Gao,
  Yu~Su, and Huan Sun.
\newblock {ScienceAgentBench: Toward Rigorous Assessment of Language Agents for
  Data-Driven Scientific Discovery}, 2024.
\newblock arXiv:2410.05080 [cs.CL].

\bibitem[Lewis et~al.(2020)Lewis, Perez, Piktus, Petroni, Karpukhin, Goyal,
  Küttler, Lewis, tau Yih, Rocktäschel, Riedel, and Kiela]{lewis2020rag}
Patrick Lewis, Ethan Perez, Aleksandra Piktus, Fabio Petroni, Vladimir
  Karpukhin, Naman Goyal, Heinrich Küttler, Mike Lewis, Wen tau Yih, Tim
  Rocktäschel, Sebastian Riedel, and Douwe Kiela.
\newblock {Retrieval-Augmented Generation for Knowledge-Intensive NLP Tasks},
  2020.
\newblock arXiv:2005.11401 [cs.CL].

\bibitem[Schick et~al.(2023)Schick, Dwivedi-Yu, Dessì, Raileanu, Lomeli,
  Zettlemoyer, Cancedda, and Scialom]{schick2023toolformer}
Timo Schick, Jane Dwivedi-Yu, Roberto Dessì, Roberta Raileanu, Maria Lomeli,
  Luke Zettlemoyer, Nicola Cancedda, and Thomas Scialom.
\newblock {Toolformer: Language Models Can Teach Themselves to Use Tools},
  2023.
\newblock arXiv:2302.04761 [cs.CL].

\bibitem[Yao et~al.(2022)Yao, Zhao, Yu, Du, Shafran, Narasimhan, and
  Cao]{yao2023react}
Shunyu Yao, Jeffrey Zhao, Dian Yu, Nan Du, Izhak Shafran, Karthik Narasimhan,
  and Yuan Cao.
\newblock {ReAct: Synergizing Reasoning and Acting in Language Models}, 2022.
\newblock arXiv:2210.03629 [cs.CL].

\bibitem[Wang et~al.(2023)Wang, Xie, Jiang, Mandlekar, Xiao, Zhu, Fan, and
  Anandkumar]{wang2023voyager}
Guanzhi Wang, Yuqi Xie, Yunfan Jiang, Ajay Mandlekar, Chaowei Xiao, Yuke Zhu,
  Linxi Fan, and Anima Anandkumar.
\newblock {Voyager: An Open-Ended Embodied Agent with Large Language Models},
  2023.
\newblock arXiv:2305.16291 [cs.AI].

\bibitem[Liu et~al.(2024)Liu, Lin, Hewitt, Paranjape, Bevilacqua, Petroni, and
  Liang]{liu2024lost}
Nelson~F. Liu, Kevin Lin, John Hewitt, Ashwin Paranjape, Michele Bevilacqua,
  Fabio Petroni, and Percy Liang.
\newblock {Lost in the Middle: How Language Models Use Long Contexts}.
\newblock \emph{Transactions of the Association for Computational Linguistics},
  12:\penalty0 157–173, 2024.
\newblock ISSN 2307-387X.
\newblock \doi{10.1162/tacl_a_00638}.

\bibitem[Shen et~al.(2026)Shen, Cheng, Ma, Turcan, Zhang, and
  Ma]{shen2026skillfoundry}
Shuaike Shen, Wenduo Cheng, Mingqian Ma, Alistair Turcan, Martin~Jinye Zhang,
  and Jian Ma.
\newblock {SKILLFOUNDRY: Building Self-Evolving Agent Skill Libraries from
  Heterogeneous Scientific Resources}, 2026.
\newblock arXiv:2604.03964 [cs.AI].

\bibitem[Li et~al.(2026{\natexlab{a}})Li, Liu, Chen, You, Di, He, Zheng, Choe,
  Sun, Wang, Tao, Li, Zhao, Geng, Wu, Zhou, Chen, Xing, Li, Zeng, Wang, Wang,
  Chaim, Jiang, Shen, Kong, Liu, Wang, Liu, Li, Lan, Lin, Ye, He, Li, Zhang,
  Gao, Li, Ma, Jing, Wang, Li, Xue, Lyu, He, Tian, Wu, Wang, Gao, Chen, Liu,
  Cheng, Bao, Tong, Xu, Zhuo, Ye, Qi, Li, Liao, Tan, Shi, Tang, Tankasala,
  Yuan, Qian, Tu, Wang, Sun, Wang, Taylor, Yang, Guan, Dong, Zhang, Dillmann,
  chung Lee, and Song]{li2026skillsbench}
Xiangyi Li, Yimin Liu, Wenbo Chen, Bingran You, Zonglin Di, Yifeng He, Shenghan
  Zheng, Kyoung~Whan Choe, Jiankai Sun, Shuyi Wang, Chujun Tao, Binxu Li,
  Xuandong Zhao, Hejia Geng, Xiaojun Wu, Junwei Zhou, Xiaokun Chen, Hanwen
  Xing, Yubo Li, Qunhong Zeng, Di~Wang, Yuanli Wang, Roey~Ben Chaim, Penghao
  Jiang, Haotian Shen, Luyang Kong, Xinyi Liu, Runhui Wang, Xuanqing Liu,
  Jiachen Li, Xin Lan, Yueqian Lin, Wengao Ye, Junwei He, Songlin Li, Yue
  Zhang, Yipeng Gao, Yijiang Li, Ze~Ma, Liqiang Jing, Tianyu Wang, Kaixin Li,
  Yiqi Xue, Haoran Lyu, Yizhuo He, Yuchen Tian, Shutong Wu, Bowei Wang, Yixuan
  Gao, Bo~Chen, Litong Liu, Sikai Cheng, Jiajun Bao, Shuaicheng Tong, Shuwen
  Xu, Terry~Yue Zhuo, Tinghan Ye, Qi~Qi, Miao Li, Longtai Liao, Zelin Tan,
  Chang Shi, Xilin Tang, Srinath Tankasala, Boqin Yuan, Yaoyao Qian, Jianhong
  Tu, Chenguang Wang, Yizhou Sun, Wei Wang, Aaron Taylor, Ziyue Yang, Changkun
  Guan, Zhikang Dong, Xinyu Zhang, Steven Dillmann, Han chung Lee, and Dawn
  Song.
\newblock {SkillsBench: Benchmarking How Well Agent Skills Work Across Diverse
  Tasks}, 2026{\natexlab{a}}.
\newblock arXiv:2602.12670 [cs.AI].

\bibitem[Ling et~al.(2026)Ling, Zhong, and Huang]{ling2026datadriven}
George Ling, Shanshan Zhong, and Richard Huang.
\newblock {Agent Skills: A Data-Driven Analysis of Claude Skills for Extending
  Large Language Model Functionality}, 2026.
\newblock arXiv:2602.08004 [cs.SE].

\bibitem[Li(2026)]{li2026dynamic}
Yubo Li.
\newblock {Dynamic Agent Skills: A Lifecycle Survey and Taxonomy of Evolving
  Skill Libraries}, 2026.
\newblock arXiv:2607.10113 [cs.AI].

\bibitem[Liu et~al.(2026{\natexlab{a}})Liu, Ji, An, Jaakkola, Zhang, and
  Chang]{liu2026realistic}
Yujian Liu, Jiabao Ji, Li~An, Tommi Jaakkola, Yang Zhang, and Shiyu Chang.
\newblock {How Well Do Agentic Skills Work in the Wild: Benchmarking LLM Skill
  Usage in Realistic Settings}, 2026{\natexlab{a}}.
\newblock arXiv:2604.04323 [cs.CL].

\bibitem[Chacko et~al.(2026)Chacko, Hugglestone, Islam, and
  Liu]{chacko2026negative}
Samuel~Jacob Chacko, James Hugglestone, Chashi~Mahiul Islam, and Xiuwen Liu.
\newblock {When Skills Don't Help: A Negative Result on Procedural Knowledge
  for Tool-Grounded Agents in Offensive Cybersecurity}, 2026.
\newblock arXiv:2605.20023 [cs.AI].

\bibitem[Greshake et~al.(2023)Greshake, Abdelnabi, Mishra, Endres, Holz, and
  Fritz]{greshake2023injection}
Kai Greshake, Sahar Abdelnabi, Shailesh Mishra, Christoph Endres, Thorsten
  Holz, and Mario Fritz.
\newblock {Not What You’ve Signed Up For: Compromising Real-World
  LLM-Integrated Applications with Indirect Prompt Injection}.
\newblock In \emph{Proceedings of the 16th ACM Workshop on Artificial
  Intelligence and Security}, CCS ’23, page 79–90. ACM, November 2023.
\newblock \doi{10.1145/3605764.3623985}.

\bibitem[Jiang et~al.(2026)Jiang, Li, Deng, Ma, Wang, Wang, and
  Yu]{jiang2026sok}
Yanna Jiang, Delong Li, Haiyu Deng, Baihe Ma, Xu~Wang, Qin Wang, and Guangsheng
  Yu.
\newblock {SoK: Agentic Skills -- Beyond Tool Use in LLM Agents}, 2026.
\newblock arXiv:2602.20867 [cs.CR].

\bibitem[Liu et~al.(2026{\natexlab{b}})Liu, Wang, Feng, Zhang, Xu, Deng, Li,
  and Zhang]{liu2026wild}
Yi~Liu, Weizhe Wang, Ruitao Feng, Yao Zhang, Guangquan Xu, Gelei Deng, Yuekang
  Li, and Leo Zhang.
\newblock {Agent Skills in the Wild: An Empirical Study of Security
  Vulnerabilities at Scale}, 2026{\natexlab{b}}.
\newblock arXiv:2601.10338 [cs.CR].

\bibitem[Li et~al.(2026{\natexlab{b}})Li, Wu, Ling, Cui, and Luo]{li2026secure}
Zhiyuan Li, Jingzheng Wu, Xiang Ling, Xing Cui, and Tianyue Luo.
\newblock {Towards Secure Agent Skills: Architecture, Threat Taxonomy, and
  Security Analysis}, 2026{\natexlab{b}}.
\newblock arXiv:2604.02837 [cs.CR].

\bibitem[{Cisco AI Defense}(2026)]{cisco2026skillscanner}
{Cisco AI Defense}.
\newblock Skill scanner.
\newblock \url{https://github.com/cisco-ai-defense/skill-scanner}, 2026.
\newblock Version 2.0.13; accessed 2026-09-02.

\bibitem[Holzbauer et~al.(2026)Holzbauer, Schmidt, Gegenhuber, Schrittwieser,
  and Ullrich]{holzbauer2026context}
Florian Holzbauer, David Schmidt, Gabriel Gegenhuber, Sebastian Schrittwieser,
  and Johanna Ullrich.
\newblock {Context Matters: Repository-Aware Security Analysis of the Agent
  Skill Ecosystem}, 2026.
\newblock arXiv:2603.16572 [cs.CR].

\bibitem[Wolf et~al.(2018)Wolf, Angerer, and Theis]{wolf2018scanpy}
F.~Alexander Wolf, Philipp Angerer, and Fabian~J. Theis.
\newblock {SCANPY: large-scale single-cell gene expression data analysis}.
\newblock \emph{Genome Biology}, 19\penalty0 (1):\penalty0 15, February 2018.
\newblock ISSN 1474-760X.
\newblock \doi{10.1186/s13059-017-1382-0}.

\bibitem[Muzellec et~al.(2023)Muzellec, Teleńczuk, Cabeli, and
  Andreux]{muzellec2023pydeseq2}
Boris Muzellec, Maria Teleńczuk, Vincent Cabeli, and Mathieu Andreux.
\newblock {PyDESeq2: a python package for bulk RNA-seq differential expression
  analysis}.
\newblock \emph{Bioinformatics}, 39\penalty0 (9):\penalty0 btad547, September
  2023.
\newblock ISSN 1367-4811.
\newblock \doi{10.1093/bioinformatics/btad547}.

\bibitem[Button et~al.(2013)Button, Ioannidis, Mokrysz, Nosek, Flint, Robinson,
  and Munafò]{button2013power}
Katherine~S. Button, John P.~A. Ioannidis, Claire Mokrysz, Brian~A. Nosek,
  Jonathan Flint, Emma S.~J. Robinson, and Marcus~R. Munafò.
\newblock {Power failure: why small sample size undermines the reliability of
  neuroscience}.
\newblock \emph{Nature Reviews Neuroscience}, 14\penalty0 (5):\penalty0
  365–376, April 2013.
\newblock ISSN 1471-0048.
\newblock \doi{10.1038/nrn3475}.

\bibitem[Wasserstein and Lazar(2016)]{wasserstein2016asa}
Ronald~L. Wasserstein and Nicole~A. Lazar.
\newblock {The ASA Statement on p-Values: Context, Process, and Purpose}.
\newblock \emph{The American Statistician}, 70\penalty0 (2):\penalty0
  129–133, April 2016.
\newblock ISSN 1537-2731.
\newblock \doi{10.1080/00031305.2016.1154108}.

\bibitem[Shaposhnikov et~al.(2026)Shaposhnikov, Fortuin, Stipcich, Gorinova,
  Heineike, and Willoughby]{shaposhnikov2026evaluation}
Maksim Shaposhnikov, Nicolas Fortuin, Simon Stipcich, Maria~I. Gorinova, Amy
  Heineike, and Rob Willoughby.
\newblock {A Framework for Evaluating Agentic Skills at Scale}, 2026.
\newblock arXiv:2606.17819 [cs.SE].

\bibitem[Kassis(2026)]{kassis2026mimeo}
Timothy Kassis.
\newblock {mimeo: Compiling Public Expert Corpora into Agent Skills and Testing
  What Transfers}, 2026.
\newblock arXiv:2609.00453 [cs.AI].

\bibitem[Harris et~al.(2020)Harris, Millman, van~der Walt, Gommers, Virtanen,
  Cournapeau, Wieser, Taylor, Berg, Smith, Kern, Picus, Hoyer, van Kerkwijk,
  Brett, Haldane, del Río, Wiebe, Peterson, Gérard-Marchant, Sheppard, Reddy,
  Weckesser, Abbasi, Gohlke, and Oliphant]{harris2020array}
Charles~R. Harris, K.~Jarrod Millman, Stéfan~J. van~der Walt, Ralf Gommers,
  Pauli Virtanen, David Cournapeau, Eric Wieser, Julian Taylor, Sebastian Berg,
  Nathaniel~J. Smith, Robert Kern, Matti Picus, Stephan Hoyer, Marten~H. van
  Kerkwijk, Matthew Brett, Allan Haldane, Jaime~Fernández del Río, Mark
  Wiebe, Pearu Peterson, Pierre Gérard-Marchant, Kevin Sheppard, Tyler Reddy,
  Warren Weckesser, Hameer Abbasi, Christoph Gohlke, and Travis~E. Oliphant.
\newblock {Array programming with NumPy}.
\newblock \emph{Nature}, 585\penalty0 (7825):\penalty0 357–362, September
  2020.
\newblock ISSN 1476-4687.
\newblock \doi{10.1038/s41586-020-2649-2}.

\bibitem[Hunter(2007)]{hunter2007matplotlib}
John~D. Hunter.
\newblock {Matplotlib: A 2D Graphics Environment}.
\newblock \emph{Computing in Science \& Engineering}, 9\penalty0 (3):\penalty0
  90–95, 2007.
\newblock ISSN 1521-9615.
\newblock \doi{10.1109/mcse.2007.55}.

\bibitem[Di~Tommaso et~al.(2017)Di~Tommaso, Chatzou, Floden, Barja, Palumbo,
  and Notredame]{ditommaso2017nextflow}
Paolo Di~Tommaso, Maria Chatzou, Evan~W Floden, Pablo~Prieto Barja, Emilio
  Palumbo, and Cedric Notredame.
\newblock {Nextflow enables reproducible computational workflows}.
\newblock \emph{Nature Biotechnology}, 35\penalty0 (4):\penalty0 316–319,
  April 2017.
\newblock ISSN 1546-1696.
\newblock \doi{10.1038/nbt.3820}.

\bibitem[Köster and Rahmann(2012)]{koster2012snakemake}
Johannes Köster and Sven Rahmann.
\newblock {Snakemake—a scalable bioinformatics workflow engine}.
\newblock \emph{Bioinformatics}, 28\penalty0 (19):\penalty0 2520–2522, August
  2012.
\newblock ISSN 1367-4803.
\newblock \doi{10.1093/bioinformatics/bts480}.

\bibitem[Grüning et~al.(2018)Grüning, Dale, Sjödin, Chapman, Rowe,
  Tomkins-Tinch, Valieris, and Köster]{gruening2018bioconda}
Björn Grüning, Ryan Dale, Andreas Sjödin, Brad~A. Chapman, Jillian Rowe,
  Christopher~H. Tomkins-Tinch, Renan Valieris, and Johannes Köster.
\newblock {Bioconda: sustainable and comprehensive software distribution for
  the life sciences}.
\newblock \emph{Nature Methods}, 15\penalty0 (7):\penalty0 475–476, July
  2018.
\newblock ISSN 1548-7105.
\newblock \doi{10.1038/s41592-018-0046-7}.

\bibitem[Wilkinson et~al.(2016)Wilkinson, Dumontier, Aalbersberg, Appleton,
  Axton, Baak, Blomberg, Boiten, da~Silva~Santos, Bourne, Bouwman, Brookes,
  Clark, Crosas, Dillo, Dumon, Edmunds, Evelo, Finkers, Gonzalez-Beltran, Gray,
  Groth, Goble, Grethe, Heringa, ’t Hoen, Hooft, Kuhn, Kok, Kok, Lusher,
  Martone, Mons, Packer, Persson, Rocca-Serra, Roos, van Schaik, Sansone,
  Schultes, Sengstag, Slater, Strawn, Swertz, Thompson, van~der Lei, van
  Mulligen, Velterop, Waagmeester, Wittenburg, Wolstencroft, Zhao, and
  Mons]{wilkinson2016fair}
Mark~D. Wilkinson, Michel Dumontier, IJsbrand~Jan Aalbersberg, Gabrielle
  Appleton, Myles Axton, Arie Baak, Niklas Blomberg, Jan-Willem Boiten,
  Luiz~Bonino da~Silva~Santos, Philip~E. Bourne, Jildau Bouwman, Anthony~J.
  Brookes, Tim Clark, Mercè Crosas, Ingrid Dillo, Olivier Dumon, Scott
  Edmunds, Chris~T. Evelo, Richard Finkers, Alejandra Gonzalez-Beltran,
  Alasdair~J.G. Gray, Paul Groth, Carole Goble, Jeffrey~S. Grethe, Jaap
  Heringa, Peter~A.C ’t Hoen, Rob Hooft, Tobias Kuhn, Ruben Kok, Joost Kok,
  Scott~J. Lusher, Maryann~E. Martone, Albert Mons, Abel~L. Packer, Bengt
  Persson, Philippe Rocca-Serra, Marco Roos, Rene van Schaik, Susanna-Assunta
  Sansone, Erik Schultes, Thierry Sengstag, Ted Slater, George Strawn,
  Morris~A. Swertz, Mark Thompson, Johan van~der Lei, Erik van Mulligen, Jan
  Velterop, Andra Waagmeester, Peter Wittenburg, Katherine Wolstencroft, Jun
  Zhao, and Barend Mons.
\newblock {The FAIR Guiding Principles for scientific data management and
  stewardship}.
\newblock \emph{Scientific Data}, 3\penalty0 (1):\penalty0 160018, March 2016.
\newblock ISSN 2052-4463.
\newblock \doi{10.1038/sdata.2016.18}.

\bibitem[Smith et~al.(2016)Smith, Katz, and Niemeyer]{smith2016software}
Arfon~M. Smith, Daniel~S. Katz, and Kyle~E. Niemeyer.
\newblock {Software citation principles}.
\newblock \emph{PeerJ Computer Science}, 2:\penalty0 e86, September 2016.
\newblock ISSN 2376-5992.
\newblock \doi{10.7717/peerj-cs.86}.

\bibitem[Wilson et~al.(2017)Wilson, Bryan, Cranston, Kitzes, Nederbragt, and
  Teal]{wilson2017goodenough}
Greg Wilson, Jennifer Bryan, Karen Cranston, Justin Kitzes, Lex Nederbragt, and
  Tracy~K. Teal.
\newblock {Good enough practices in scientific computing}.
\newblock \emph{PLOS Computational Biology}, 13\penalty0 (6):\penalty0
  e1005510, June 2017.
\newblock ISSN 1553-7358.
\newblock \doi{10.1371/journal.pcbi.1005510}.

\bibitem[Liu et~al.(2023)Liu, Xia, Wang, and Zhang]{liu2023evalplus}
Jiawei Liu, Chunqiu~Steven Xia, Yuyao Wang, and Lingming Zhang.
\newblock {Is Your Code Generated by ChatGPT Really Correct? Rigorous
  Evaluation of Large Language Models for Code Generation}, 2023.
\newblock arXiv:2305.01210 [cs.SE].

\bibitem[Agent Plugins()]{agentplugins2026spec}
Agent Plugins.
\newblock Agent plugins 1.0.0 specification.
\newblock \url{https://agent-plugins.org/}, 2026.
\newblock Accessed 2026-09-02.

\bibitem[Cock et~al.(2009)Cock, Antao, Chang, Chapman, Cox, Dalke, Friedberg,
  Hamelryck, Kauff, Wilczynski, and de~Hoon]{cock2009biopython}
Peter J.~A. Cock, Tiago Antao, Jeffrey~T. Chang, Brad~A. Chapman, Cymon~J. Cox,
  Andrew Dalke, Iddo Friedberg, Thomas Hamelryck, Frank Kauff, Bartek
  Wilczynski, and Michiel J.~L. de~Hoon.
\newblock {Biopython: freely available Python tools for computational molecular
  biology and bioinformatics}.
\newblock \emph{Bioinformatics}, 25\penalty0 (11):\penalty0 1422–1423, March
  2009.
\newblock ISSN 1367-4803.
\newblock \doi{10.1093/bioinformatics/btp163}.

\bibitem[Li et~al.(2009)Li, Handsaker, Wysoker, Fennell, Ruan, Homer, Marth,
  Abecasis, and Durbin]{li2009sequence}
Heng Li, Bob Handsaker, Alec Wysoker, Tim Fennell, Jue Ruan, Nils Homer, Gabor
  Marth, Goncalo Abecasis, and Richard Durbin.
\newblock {The Sequence Alignment/Map format and SAMtools}.
\newblock \emph{Bioinformatics}, 25\penalty0 (16):\penalty0 2078–2079, June
  2009.
\newblock ISSN 1367-4803.
\newblock \doi{10.1093/bioinformatics/btp352}.

\bibitem[RDKit()]{rdkit}
RDKit.
\newblock {RDKit: Open-Source Cheminformatics}.
\newblock \url{https://www.rdkit.org}, 2026.
\newblock Accessed 2026-09-02.

\bibitem[Love et~al.(2014)Love, Huber, and Anders]{love2014moderated}
Michael~I Love, Wolfgang Huber, and Simon Anders.
\newblock {Moderated estimation of fold change and dispersion for RNA-seq data
  with DESeq2}.
\newblock \emph{Genome Biology}, 15\penalty0 (12):\penalty0 550, December 2014.
\newblock ISSN 1474-760X.
\newblock \doi{10.1186/s13059-014-0550-8}.

\bibitem[Schmotz et~al.(2025)Schmotz, Abdelnabi, and
  Andriushchenko]{schmotz2025injection}
David Schmotz, Sahar Abdelnabi, and Maksym Andriushchenko.
\newblock {Agent Skills Enable a New Class of Realistic and Trivially Simple
  Prompt Injections}, 2025.
\newblock arXiv:2510.26328 [cs.LG].

\bibitem[Debenedetti et~al.(2024)Debenedetti, Zhang, Balunović,
  Beurer-Kellner, Fischer, and Tramèr]{debenedetti2024agentdojo}
Edoardo Debenedetti, Jie Zhang, Mislav Balunović, Luca Beurer-Kellner, Marc
  Fischer, and Florian Tramèr.
\newblock {AgentDojo: A Dynamic Environment to Evaluate Prompt Injection
  Attacks and Defenses for LLM Agents}, 2024.
\newblock arXiv:2406.13352 [cs.CR].

\end{thebibliography}
